\documentclass[10pt,journal,compsoc]{IEEEtran}

\usepackage[linesnumbered,ruled]{algorithm2e}
\usepackage{algpseudocode}
\usepackage{times}
\usepackage{epsfig}
\usepackage{graphicx}
\usepackage{color}
\usepackage{graphics} 
\usepackage{epsfig} 
\usepackage{amsmath} 
\usepackage{amssymb}  
\usepackage{epstopdf}
\usepackage{mathtools}
\usepackage{stfloats}
\usepackage{makecell}
\usepackage{multirow}
\usepackage{pifont}
\DeclareMathOperator*{\argmax}{argmax}

\newcommand{\PreserveBackslash}[1]{\let\temp=\\#1\let\\=\temp}

\newcolumntype{C}[1]{>{\PreserveBackslash\centering}p{#1}}
\newcolumntype{R}[1]{>{\PreserveBackslash\raggedleft}p{#1}}
\newcolumntype{L}[1]{>{\PreserveBackslash\raggedright}p{#1}}

\usepackage[pagebackref=false,breaklinks=true,letterpaper=true,colorlinks,bookmarks=false,citecolor=blue]{hyperref}
\usepackage{cite}

\def \TRK{\mathcal{T}}
\def \VRF{\mathcal{V}}

\def \TK{\mathcal{T}}
\def \VF{\mathcal{V}}

\def\c{\textbf{c}}

\def\h{\textbf{h}}

\def\w{\textbf{w}}
\def\x{\textbf{x}}

\def\z{\textbf{z}}

\ifCLASSINFOpdf

\else

\fi

\begin{document}

\title{Parallel Tracking and Verifying}

\author{Heng~Fan
        and~Haibin~Ling
}

%
%

\markboth{Draft}%
{Shell Submitted to IEEE Transactions on Pattern Analysis and Machine Intelligence}
%

\IEEEtitleabstractindextext{%
\begin{abstract}
Being intensively studied, visual object tracking has witnessed great advances in either speed (e.g., with correlation filters) or accuracy (e.g., with deep features). Real-time and high accuracy tracking algorithms, however, remain scarce. In this paper we study the problem from a new perspective and present a novel \emph{parallel tracking and verifying} (PTAV) framework, by taking advantage of the ubiquity of multi-thread techniques and borrowing ideas from the success of \emph{parallel tracking and mapping} in visual SLAM. The proposed PTAV framework is typically composed of two components, a (base) tracker $\TRK$ and a verifier $\VRF$, working in parallel on two separate threads. The tracker $\TRK$ aims to provide a super real-time tracking inference and is expected to perform well most of the time; by contrast, the verifier $\VRF$ validates the tracking results and corrects $\TRK$ when needed. The key innovation is that, $\VRF$ does not work on every frame but only upon the requests from $\TRK$; on the other end, $\TRK$ may adjust the tracking according to the feedback from $\VRF$. With such collaboration, PTAV enjoys both the high efficiency provided by $\TRK$ and the strong discriminative power by $\VRF$. Meanwhile, to adapt $\VRF$ to object appearance changes over time, we maintain a dynamic target template pool for adaptive verification, resulting in further performance improvements. In our extensive experiments on popular benchmarks including OTB2015, TC128, UAV20L and VOT2016, PTAV achieves the best tracking accuracy among all real-time trackers, and in fact even outperforms many deep learning based algorithms. Moreover, as a general framework, PTAV is very flexible with great potentials for future improvement and generalization.
\end{abstract}

}

\maketitle

\IEEEdisplaynontitleabstractindextext

\IEEEpeerreviewmaketitle

\section{Introduction}

\subsection{Background}

\IEEEPARstart{A}{s} one of the most important components in computer vision, visual tracking has a long list of applications such as robotics, intelligent vehicles, visual surveillance, human-computer interaction and so forth~\cite{smeulders2014visual,YilmazJS06survey,li2013survey,kristan2016novel}. Given an initial state (usually a bounding box) of a tracking target in the first frame, visual tracking aims at estimating the unknown states (e.g., position and scale) of the target object in subsequent consecutive frames. Although significant progresses have been made in recent decades, robust object tracking still remains challenging due to large appearance variations caused by many factors such as object occlusion, deformation, rotation, illumination variations, scale changes, motion blur and so on.

\begin{figure}[!t]
\centering
\includegraphics[width=\linewidth]{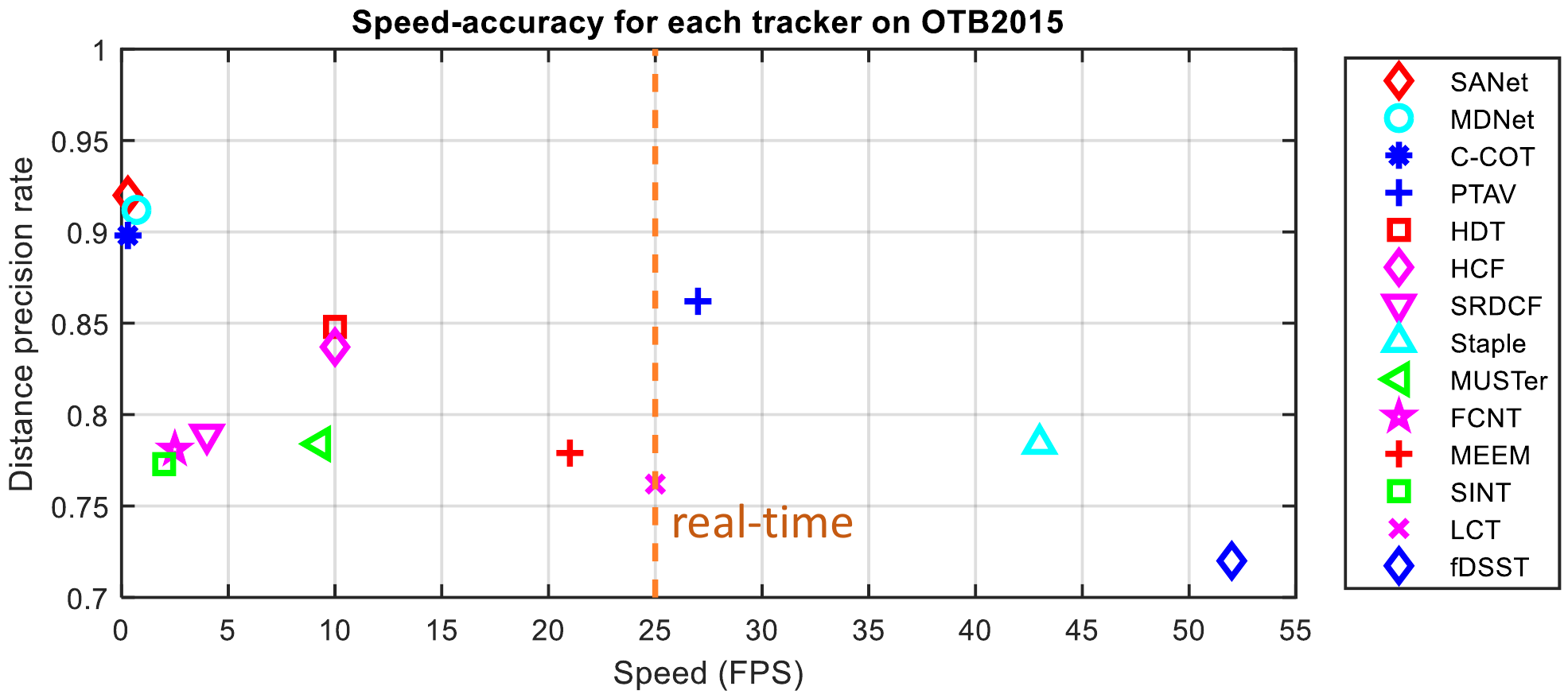}\\
\caption{Speed-accuracy plot of state-of-the-art trackers on OTB2015~\cite{wu2015object}. For better illustration, only those trackers with accuracy higher than 0.7 are reported. Compared with high precision deep learning-based trackers (e.g., MDNet, SANet and C-COT) whose speeds are around 1 FPS, our PTAV runs in real-time without serious accuracy degradation. On the other hand, compared with other real-time trackers (e.g., Staple, LCT and fDSST), PTAV achieves a much higher accuracy. Moreover, PTAV even outperforms some deep learning-based trackers in both accuracy and speed (e.g., HCF, HDT, SINT and FCNT).}
\label{fig:speed-accuracy}
\end{figure}

Recently, inspired by the success of deep convolutional neural networks (CNNs)~\cite{lecun1989backpropagation} in image recognition (e.g.,~\cite{krizhevsky2012imagenet}), an emerging trend toward improving tracking accuracy is to utilize robust deep features for object appearance representation (e.g.,~\cite{ma2015hierarchical,qi2016hedged,nam2016learning,wang2016stct,wang2013learning,fan2016sanet,danelljan2016beyond,tao2016siamese,wang2015visual,hong2015online}). Despite significant improvements obtained in accuracy, these algorithms often suffer from high computational burden due to either extracting expensive deep features (e.g.,~\cite{ma2015hierarchical,qi2016hedged,danelljan2016beyond,tao2016siamese,wang2015visual,hong2015online}) or online network fine-tuning (e.g.,~\cite{nam2016learning,wang2016stct,wang2013learning,fan2016sanet}), and hardly meet the real-time requirement (see Figure~\ref{fig:speed-accuracy} for illustration).

Along a somewhat orthogonal direction, researchers have been proposing efficient visual trackers (e.g., \cite{henriques2015high,zhang2012real,danelljan2016discriminative,bolme2010visual,bertinetto2016staple,henriques2012exploiting,danelljan2014adaptive,kalal2012tracking}), notably represented by the series of trackers based on correlation filters. To achieve efficient computation, the correlation filter-based trackers usually represent object appearance with simple hand-crafted features such as raw pixels, HoG~\cite{dalal2005histograms} and color names~\cite{van2009learning}. While easily running at real-time, these trackers usually perform less robustly compared to deep learning-based approaches (see again Figure~\ref{fig:speed-accuracy}).

Despite aforementioned progresses in either speed or accuracy, real-time high quality tracking algorithms remain scarce. A natural way is to seek a trade-off between speed and accuracy (e.g.,~\cite{bertinetto2016staple,ma2015long,bertinetto2016fully}). In this paper we work toward this goal, but from a novel perspective described as the following.

\begin{figure*}[!t]
\centering
\includegraphics[width=\linewidth]{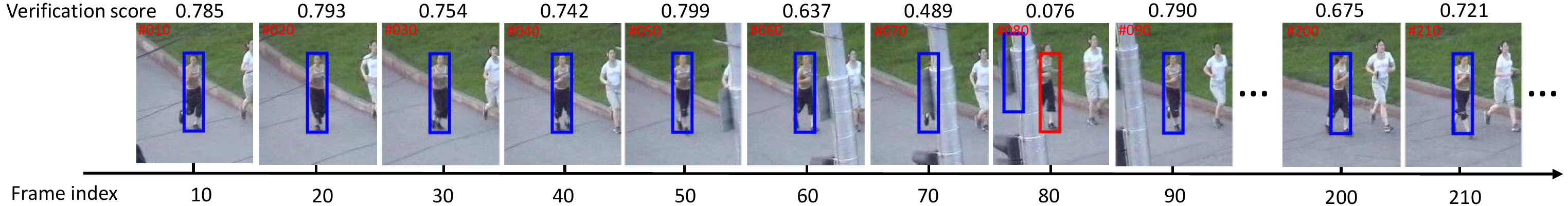}\\
\caption{Illustration of verifying scores on a typical sequence. Verifier validates tracking results every 10 frames. Most of the time the tracking results are reliable (showing in \textcolor{blue}{blue}). Occasionally, e.g., frame \#080, the verifier finds the original tracking result (showing in \textcolor{blue}{blue}) unreliable and the tracker is corrected and resumes tracking based on detection result provided by verifier (showing in \textcolor{red}{red}).}
\label{fig:10frames}
\end{figure*}

\subsection{Motivation}

Our key idea is to decompose the original tracking task into two parallel but collaborative ones, one for fast tracking and the other for accurate verification. We are mainly inspired by the following observations or related works:
\\
\textbf{Motivation 1.} When tracking a target from visual input, most of the time the target object moves smoothly and its appearance changes slowly or remains the same. Simple but efficient algorithms usually work fine for such easy cases. By contrast, hard cases (e.g., drastic object appearance variations) happen only occasionally, though they can cause serious consequences if not addressed properly. These hard cases usually require computationally expensive processes or analysis, such as the verification in our approach. Intuitively, verifications are needed only occasionally instead of for every frame. Figure~\ref{fig:10frames} shows a typical example with both cases.
\\
\textbf{Motivation 2.} The ubiquity of multi-thread computing has already benefited computer vision systems, with notably in visual SLAM (\emph{simultaneous localization and mapping}). By splitting tracking and mapping into two parallel threads, PTAM (\emph{parallel tracking and mapping})~\cite{klein2007parallel} provides one of the most popular SLAM frameworks with many important extensions~(e.g., ORB-SLAM~\cite{ORB-SLAM}). A key inspiration in PTAM is that mapping is not needed for every frame. Nor does verifying in our task.
\\
\textbf{Motivation 3.} Last but not least, recent advances in either fast or accurate tracking algorithms provide promising building blocks and highly encourage us to seek a balanced system for real-time high accuracy visual tracking.

\subsection{Contribution}

With the motivations listed above, we propose to build real-time high accuracy trackers in a novel framework named \emph{parallel tracking and verifying} (PTAV). PTAV typically consists of two components: a fast tracker\footnote{For conciseness, in the rest of this paper, we refer the \emph{fast tracker} as a \emph{tracker}, whenever no confusion caused.} denoted by $\TRK$ and an accurate verifier denoted by $\VRF$.  The two components work in parallel on two separate threads while collaborating with each other. The tracker $\TRK$ aims at providing a super real-time tracking inference and is expected to perform well most of the time, e.g., most frames in Figure~\ref{fig:10frames}. By contrast, the verifier $\VRF$ checks the tracking results and corrects $\TRK$ when needed, e.g., at frame \#080 in Figure~\ref{fig:10frames}.

The key idea is, while $\TRK$ needs to run on every frame, $\VRF$ does not. As a general framework, PTAV allows the coordination between the tracker and the verifier: $\VRF$ checks the tracking results provided by $\TRK$ and sends feedback to $\VRF$; and $\VRF$ adjusts itself according to the feedback when necessary. By running $\TRK$ and $\VRF$ in parallel, PTAV inherits both the high efficiency of $\TRK$ and the strong discriminative power of $\VRF$.

Implementing a PTAV algorithm requires three parts: a base tracker for $\TRK$, a base verifier for $\VRF$, and the coordination between them. For $\TRK$, we choose the Staple algorithm~\cite{bertinetto2016staple}, which is correlation filter-based and runs efficiently by itself. For $\VRF$, we choose the Siamese network~\cite{chopra2005learning} for verification similar to that in~\cite{tao2016siamese}. For coordination, $\TRK$ sends results to $\VRF$ at an adaptive frequency that allows enough time for verification. On the verifier side, when an unreliable result is found, $\VF$ performs detection and sends the detected result to $\TRK$ for correction. For $\VF$ to handle object appearance changes over time, we utilize $k$-means clustering to maintain a dynamic target template pool for adaptive verification, resulting in further improvements of PTAV in both accuracy and speed. 

The proposed PTAV algorithm is evaluated thoroughly on several popular benchmarks including OTB2015 \cite{wu2015object}, TC128 \cite{liang2015encoding}, UAV20L \cite{mueller2016benchmark} and VOT2016~\cite{kristan2016visual}. In these experiments, PTAV achieves the best tracking accuracy among all real-time trackers, and in fact performs even better than many deep learning-based solutions.

In summary, our first main contribution is the novel parallel tracking and verifying framework. With the framework, we make the second contribution by implementing a tracking solution that combines correlation kernel-based tracking and deep learning-based verification. Then, our solution demonstrates very promising results on thorough experiments in comparison with state-of-the-arts. Moreover, it is worth noting that PTAV is a very flexible framework and our implementation may not be optimal. We believe there are great rooms for future improvement and generalization.

This paper is an extended version of a preliminary conference publication~\cite{fan2017parallel}. The main new contributions or differences include: (1) a more robust base tracker (i.e., Staple) for $\TRK$ in implementing PTAV, which brings clear performance improvement, (2) a dynamic target template pool for adaptive verification against target appearance variations, (3) various ablation studies on $\VF$ and $\TRK$ to analyze PTAV, including two base verifiers (VGGNet~\cite{simonyan2014very} and AlexNet~\cite{krizhevsky2012imagenet}) for $\VF$ and three base trackers (KCF~\cite{henriques2015high}, fDSST~\cite{danelljan2016discriminative} and Staple~\cite{bertinetto2016staple}) for $\TRK$, and (4) more thorough experimental validation and analysis involving more state-of-the-art tracking algorithms and benchmarks.

The rest of this paper starts with an extensive overview of related work in Section \ref{related_work}. Then, Section \ref{ptav} elaborates the proposed PTAV framework and implementation details. Section \ref{exp} shows experimental results, including comparisons with state-of-the-arts and ablation studies, followed by conclusion in Section \ref{con}.

\section{Related Work}
\label{related_work}
\noindent {\bf Visual tracking algorithms.} Visual tracking has been extensively studied and it is beyond our scope to review all previous studies. Instead, in the following we sample some representative works and discuss those closely related to ours. Some comprehensive reviews on object tracking can be found in~\cite{smeulders2014visual,YilmazJS06survey,li2013survey,kristan2016novel}.

The focus of this paper is on \emph{model-free} single object tracking, for which existing algorithms are often categorized as either discriminative or generative. Discriminative algorithms usually treat tracking as a classification problem that distinguishes the target from changing background. Babenko \emph{et al.}~\cite{babenko2011robust} apply multiple instant learning (MIL) to learn a classifier based on an adaptive appearance model for object tracking. Zhang \emph{et al.}~\cite{zhang2012real} propose a compressive sensing tracker by projecting high-dimensional features to low-dimensional compressed subspace. 
Grabner \emph{et al.}~\cite{grabner2008semi} propose an online tracking algorithm via semi-supervised boosting which treats samples from the first frame as labeled and other samples as unlabeled. Hare \emph{et al.}~\cite{hare2016struck} propose to leverage a kernelized structured output support vector machine (SVM) for robust visual tracking by mitigating the effect of wrong labeling samples.

By contrast, generative algorithms usually formulate tracking as searching for regions most similar to the target. To this end, numerous object appearance modeling approaches have been proposed. In~\cite{ross2008incremental}, Ross \emph{et al.} propose an online incremental subspace learning method to adapt appearance changes for object tracking. Kwon \emph{et al.}~\cite{kwon2010visual} present a modified particle filtering framework for tracking by combining multiple observation and motion models to handle large appearance and motion variations. Mei and Ling~\cite{mei2009robust} model object appearance with sparse representation and propose the $\ell_{1}$-tracker, which is later improved via an accelerated proximal gradient algorithm~\cite{bao2012real}. In~\cite{zhang2015structural}, Zhang \emph{et al.} propose to incorporate target structure into sparse representation for robustness.

\vspace{0.5em}
\noindent {\bf Deep learning-based tracking.} Motivated by the power of deep features in visual recognition (e.g.,~\cite{krizhevsky2012imagenet,simonyan2014very}), some trackers utilize deep features for object appearance modeling, and achieve excellent performance, though typically at the cost of low running speed.
Wang \emph{et al.}~\cite{wang2013learning} introduce a stacked denoising autoencoder to learn generic image features for visual tracking. 
In~\cite{wang2015visual}, Wang \emph{et al.} present a fully convolutional neural network tracking (FCNT) algorithm by transferring pre-trained CNN features to improve tracking accuracy. Ma \emph{et al.}~\cite{ma2015hierarchical} replace HoG~\cite{dalal2005histograms} with discriminative convolutional features for correlation filter tracking, resulting in remarkable performance gains. A similar idea is present by Qi \emph{et al.}~\cite{qi2016hedged} by adaptively merging convolutional features from different layers. Hong \emph{et al.}~\cite{hong2015online} propose to learn discriminative saliency map for online object tracking using CNNs. To address the problem of lack of training samples, Wang \emph{et al.}~\cite{wang2016stct} employ intermediate features in networks to learn a robust ensemble tracker. In~\cite{nam2016learning}, Nam \emph{et al.} propose to impose multiple domain branches on a light architecture of CNNs to learn generic feature for tracking target, and then introduce an online tracking algorithm by updating network weights in each frame. In~\cite{fan2016sanet}, recurrent neural networks (RNNs) are introduced to capture internal structure of a tracking target and the generated tracking algorithm achieves promising results on several tracking benchmarks.
Though these approaches have achieved very impressive results, the heavy computation burden severely restricts their practical applications.

\begin{figure*}[!t]
\centering
\includegraphics[width=.99\linewidth]{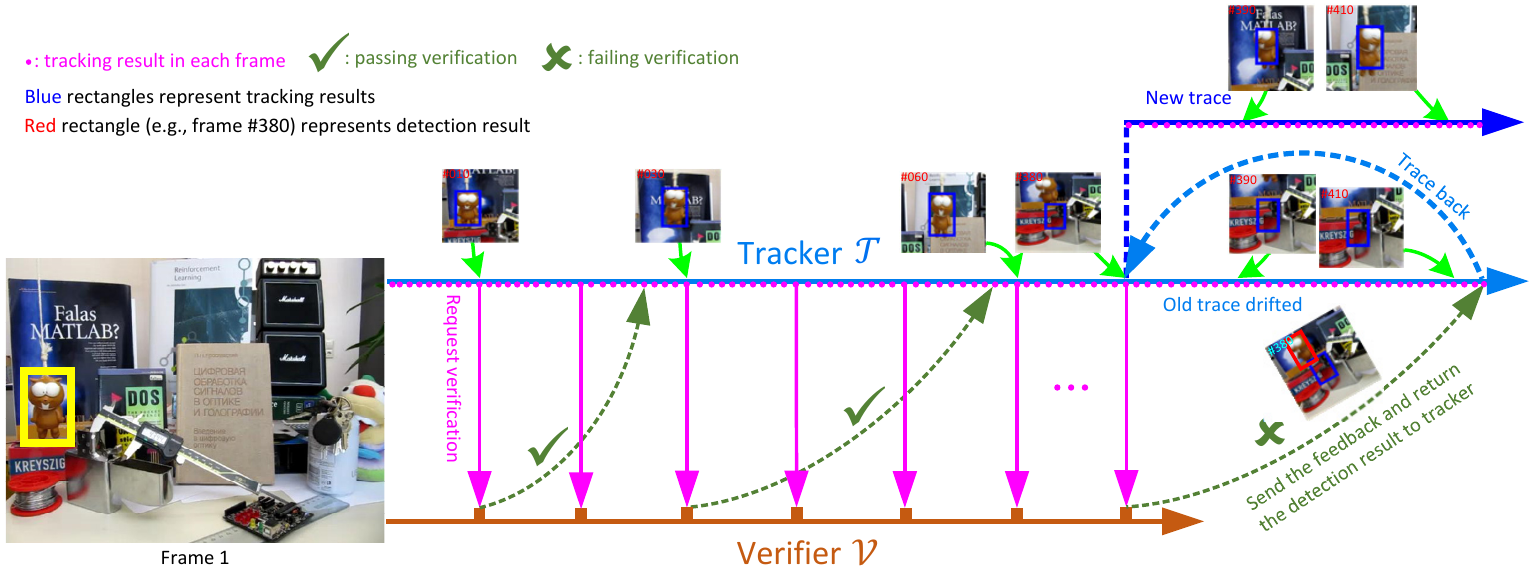}
\caption{Illustration of the PTAV framework in which tracking and verifying are processed asynchronously in two parallel threads.}
\label{detail_PTV}
\end{figure*}

\vspace{0.5em}
\noindent {\bf Correlation filter-based tracking.} Recently, correlation filter has drawn increasing attention in visual tracking owing partly to its high computation efficiency. Bolme \emph{et al.}~\cite{bolme2010visual} propose to use correlation filter for tracking through learning the minimum output sum of squared error (MOSSE). Benefitting from the high computation efficiency of correlation filter, this approach runs amazingly at hundreds of frames per second. Henriques {\it et al.}~\cite{henriques2012exploiting} incorporate kernel space into correlation filter and propose a circulant structure with kernel (CSK) method for visual object tracking, and later \cite{henriques2015high} extends CSK to the well-known kernelized correlation filters (KCF) tracker by substituting raw pixel intensities with HoG~\cite{dalal2005histograms} for appearance representation. 
To deal with the scale issue, Danelljan \emph{et al.}~\cite{danelljan2016discriminative} suggest an extra scale filter into correlation filter tracking to adaptively estimate target scale. In~\cite{li2014scale}, Li \emph{et al.} adopt a similar strategy to address the problem of scale changes.
Later, more efforts have been made to improve correlation filter tracking. Danelljan \emph{et al.}~\cite{danelljan2014adaptive} investigate color attributes to improve correlation filter tracking. To reduce the risk of model drift, Ma \emph{et al.}~\cite{ma2015long} introduce an auxiliary detector to re-locate the tracking target when sensing drift. In~\cite{liu2015real}, Liu \emph{et al.} propose a part-based correlation filter tracker by decomposing tracking target into fragments, which is robust to resist occlusion. Mueller \emph{et al.}~\cite{mueller2017context} propose to explicitly incorporate context into correlation filter to improve its discriminability. To alleviate boundary effect, the work in~\cite{danelljan2015learning} presents spatially regularized correlation filters for tracking. Bertinetto \emph{et al.}~\cite{bertinetto2016staple} introduce a complementary tracker by combing correlation filters and color histograms. In~\cite{valmadre2017end}, Valmadre \emph{et al.} propose correlation filter networks for tracking by enjoying end-to-end representation.

\vspace{0.5em}
\noindent {\bf Verification in tracking.} The idea of verification is not new for tracking. A notable example is the tracking-learning-detection (TLD) algorithm~\cite{kalal2012tracking}, in which tracking results are validated \emph{per frame} to decide how learning/detection shall progress. Similar ideas have been used in later works. In~\cite{hua2014occlusion}, Hua \emph{et al.} propose an occlusion and motion reasoning-based long-term tracker. An occlusion detector is applied in \emph{each frame} to prevent tracking model from drift. Ma \emph{et al.}~\cite{ma2015long} apply an additional objet detector in correlation filter tracking for the same end. Unlike in previous studies, the verification in PTAV runs only on sampled frames. This mechanism allows PTAV to use strong verification algorithms without worrying much about running time efficiency. In fact, we utilize the Siamese network~\cite{chopra2005learning} that is designed for verification tasks.

Interestingly, tracking by itself can be also formulated as a verification problem that finds the best candidate similar to the tracking target~\cite{tao2016siamese,bertinetto2016fully}. Bertinetto \emph{et al.}~\cite{bertinetto2016fully} propose a fully-convolutional Siamese network for visual tracking by searching the tracking target within a local region. Tao \emph{et al.}~\cite{tao2016siamese} formulate tracking problem as a task of object matching in each frame and develop a matching function based on the Siamese network. Despite obtaining excellent performance, the application of such trackers is limited by the heavy computation for extracting deep features in each frame. Compared with these studies, our solution treats verification only as a way to validate and correct the \emph{fast tracker}, and does not run verification per frame.

\vspace{0.5em}
\noindent {\bf Ensemble tracking.} To achieve robustness in tracking, a natural solution is to leverage multiple different components to determine tracking result. Kwon \emph{et al.}~\cite{kwon2010visual} combine multiple observation and motion models to handle large appearance changes in tracking. The TLD tracker~\cite{kalal2012tracking} uses both tracker and detector for long-term tracking, and a similar idea is adopted in~\cite{ma2015long}. In~\cite{bertinetto2016staple}, Bertinetto \emph{et al.} propose a complementary tracking approach based on correlation filters and color histograms. Hong \emph{et al.}~\cite{hong2015multi} present a multi-store tracker (MUSTer) which consists of short- and long-term memory stores to process target appearance. The final tracking result is jointly determined by short- and long-term memories. Different than in these studies, our PTAV is comprised of two components, which run on two parallel threads asynchronously while collaborating with each other.

Though other ensemble approaches (e.g.,~\cite{kwon2010visual,ma2015long,bertinetto2016staple,hong2015multi}) can be implemented using multiple threads as well, the proposed PTAV fundamentally differs from them. In these existing algorithms, different components are simultaneously used to determine tracking result in each frame, thus their multi-thread implementations are \emph{synchronous}. By contrast, in PTAV, $\TRK$ and $\VF$ function in different ways and run independent except for necessary interactions, and thus the multi-thread implementation is \emph{asynchronous}.

\section{Parallel Tracking and Verifying (PTAV)}
\label{ptav}
\subsection{Framework}

A typical PTAV consists of two components: a (fast) tracker $\TK$ and a (reliable) verifier $\VF$. The two components work together toward real-time and high accuracy tracking.
\begin{itemize}
\item \noindent {\bf The tracker $\TK$} is responsible of the ``real-time" requirement of PTAV, and needs to locate the target in each frame. Meanwhile, $\TK$ sends verification request to $\VF$ from time to time (though not every frame), and responds to feedback from $\VF$ by adjusting tracking or updating models. To avoid heavy computation, $\TK$ maintains a buffer of tracking information (e.g., intermediate status) in recent frames to facilitate fast tracing back when needed.

\item \noindent {\bf The verifier $\VF$} is employed to pursue the ``high accuracy" requirement of PTAV. Up on receiving a request from $\TK$, $\VF$ tries the best to first validate the tracking result (e.g., comparing it with the template), and then provide feedback to $\TK$. To adapt $\VF$ to object appearance variations over time, the tracking target template is \emph{not} fixed. Instead, $\VF$ collects a number of reliable tracking results, and then use $k$-means to cluster these results to obtain a target template pool for subsequent verification.
\end{itemize}

In PTAV, $\TK$ and $\VF$ run in parallel on two different threads with necessary interactions, as illustrated in Figure~\ref{detail_PTV}. The tracker $\TK$ and verifier $\VF$ are initialized in the first frame. After that, $\TK$ starts to process each arriving frame and generates the result (pink solid dot in Figure \ref{detail_PTV}). In the meantime, $\VF$ validates the tracking result every several frames. Because tracking is much faster than verifying, $\TK$ and $\VF$ work asynchronously. Such mechanism allows PTAV to tolerate temporary tracking drift (e.g., at frame 380 in Figure \ref{detail_PTV}), which will be corrected later by $\VF$. When $\VF$ finds a tracking result unreliable, it searches the correct answer from a local region and sends it to $\TK$. Upon the receipt of such feedback, $\TK$ stops current tracking job and traces back to resume tracking with the correction provided by $\VF$.

\begin{algorithm}[!bht]
	\caption{Parallel Tracking and Verifying (PTAV)}\label{PTAV_alg}
	Initialize the tracking thread for tracker $\TK$\;
	Initialize the verifying thread for verifier $\VF$\;
	Initialize \textit{current\_frame} as the second frame\;
	Run $\TK$ (Alg.~\ref{PTAV_T}) and $\VF$ (Alg.~\ref{PTAV_V}) till the end of tracking\;
\end{algorithm}

\begin{algorithm}[!bht]
	\caption{Tracking Thread $\TK$}\label{PTAV_T}
	\While {\textit{current\_frame} is valid}
	{
		\If {received a message from $\VF$}
		{
			\eIf{verification passed}
			{
				Update tracking model (optional)\;
			}
			{
				Correct tracking\;
				Trace back and reset \textit{current\_frame}\;
			}
		}
		Tracking on the \textit{current\_frame}\;
		\If{time for verification according to $N_\mathrm{int}$}
		{
			Send the current result to $\VF$ to verify;
		}
		\textit{current\_frame} $\leftarrow$ next frame\;
	}
\end{algorithm}

\begin{algorithm}[!bht]
	\caption{Verifying Thread $\VF$}\label{PTAV_V}
	\While{not ended}
	{
		\If {received request from $\TK$}
		{
			Verifying the tracking result\;
			Collect tracking result and perform $k$-means clustering if needed\;
			\If{verification failed}
			{
				Provide correction information in $s$\;
				Adjust $N_\mathrm{int}$ if needed\;
			}
			Send verification result $s$ to $\TK$\;
		}
	}
\end{algorithm}

It is worth noting that PTAV provides a very flexible framework, and some important designing choices are following. (1) The base algorithms for $\TK$ and $\VF$ may depend on specific applications and available computational resources. In addition, in practice one may use more than one verifiers or even base trackers. (2) The response of $\TK$ to the feedback from $\VF$, either positive or negative, can be largely designed to adjust to specific requests. (3) The correction of unreliable tracking results can be implemented in various ways, and it can even be conducted purely by $\TK$ (i.e., including target detection). (4) $\TK$ has numerous methods to use pre-computed and archived information for speeding up. Algorithms \ref{PTAV_alg}-\ref{PTAV_V} summarize the general PTAV framework.

\subsection{PTAV Implementation}

\subsubsection{Tracking}

We choose the Staple tracker~\cite{bertinetto2016staple} for $\TK$ in PTAV. The main idea of Staple is to combine two complementary cues, i.e., template and histogram, for tracking. To such end, given an image patch $\z$, a linear combination of tracking scores from template and histogram is proposed
\begin{equation}
y(\z) = (1-\alpha)y_\mathrm{tmpl}(\z)+{\alpha}y_\mathrm{hist}(\z)   \label{eq1}
\end{equation}
where $\alpha$ denotes a trade-off parameter, and $y_\mathrm{tmpl}(\z)$ and $y_\mathrm{hist}(\z)$ represent the tracking responses based on template and on histogram information, respectively.

The tracking response on template is derived by learning the optimal correlation filter model $\w$, which is efficiently solved in frequency domain through the fast Fourier transformation (FFT). At time $t$, the FFT of the filter responses is first calculated using $\w$ and an inverse FFT is then conducted to derive the final response $y_\mathrm{tmpl}(\z)$. The model $\w$ is online updated in each frame.

The tracking response on histogram is based on a learned color statistic model $\h$, which is robust in resisting deformation. At time $t$, $\h$ is utilized to calculate $y_\mathrm{hist}(\z)$, and then dynamically updated. To adapt the tracker to scale changes, a scale filter is adopted to estimate the target scale. More details about the Staple tracker can be found in~\cite{bertinetto2016staple}.

To efficiently leverage Staple as $\TK$ in PTAV, in addition to the original Staple algorithm, $\TK$ stores all intermediate results (e.g., $\w$ and $\h$) for each frame after sending out last verification request. Let $\mathcal{W}=\{\w_{\xi-\bigtriangleup},\cdots,\w_{\xi}\}$ and $\mathcal{H}=\{\h_{\xi-\bigtriangleup},\cdots,\h_{\xi}\}$ represent the collections of $\w$ and $\h$, where $\xi$ is the index of the last frame processed by $\TK$, and $\bigtriangleup$ denotes a fixed size for temporal sliding window to store tracking models. These intermediate results in $\mathcal{W}$ and $\mathcal{H}$ allow $\TK$ for fast tracing back.

In particular, when $\VRF$ detects unreliable tracking result in frame $k$ while $\TK$ starts working on frame $j$ ($j>k$), a feedback consisting of correct target position and frame index information is sent to $\TK$. Once receiving this feedback from $\VRF$, $\TK$ first stops processing frame $j$ and then utilizes the archived target position and tracking model (i.e., $\w_{k-1}$ and $\h_{k-1}$) retrieved from $\mathcal{W}$ and $\mathcal{H}$ to resume subsequent tracking from frame $k$. Meanwhile, useless intermediate results in $\mathcal{W}$ (i.e., $\w_k$ to $\w_{j-1}$) and $\mathcal{H}$ (i.e., $\h_k$ to $\h_{j-1}$) will be discarded.

Note that we do not assume the correctness of $\w_{k-1}$ and $\h_{k-1}$ in the above strategy. In fact, one way is to trace backward from $k-1$ to locate a reliable frame to resume tracking, at additional expense of more verification operations. In practice, however, we found that $\w_{k-1}$ and $\h_{k-1}$ typically provide sufficient initial guess for frame $k$ and rely on the detection part to correct the incorrect tracking result. More details are given the following sections on verifying and detection. 


To validate the tracking result, $\TK$ sends the verification results every $N_{\mathrm{int}}$ frames, where $N_{\mathrm{int}}$ denotes the dynamically adjustable verification interval as described later.

\begin{figure}[!t]
\centering
\includegraphics[width=\linewidth,height=1.25\linewidth]{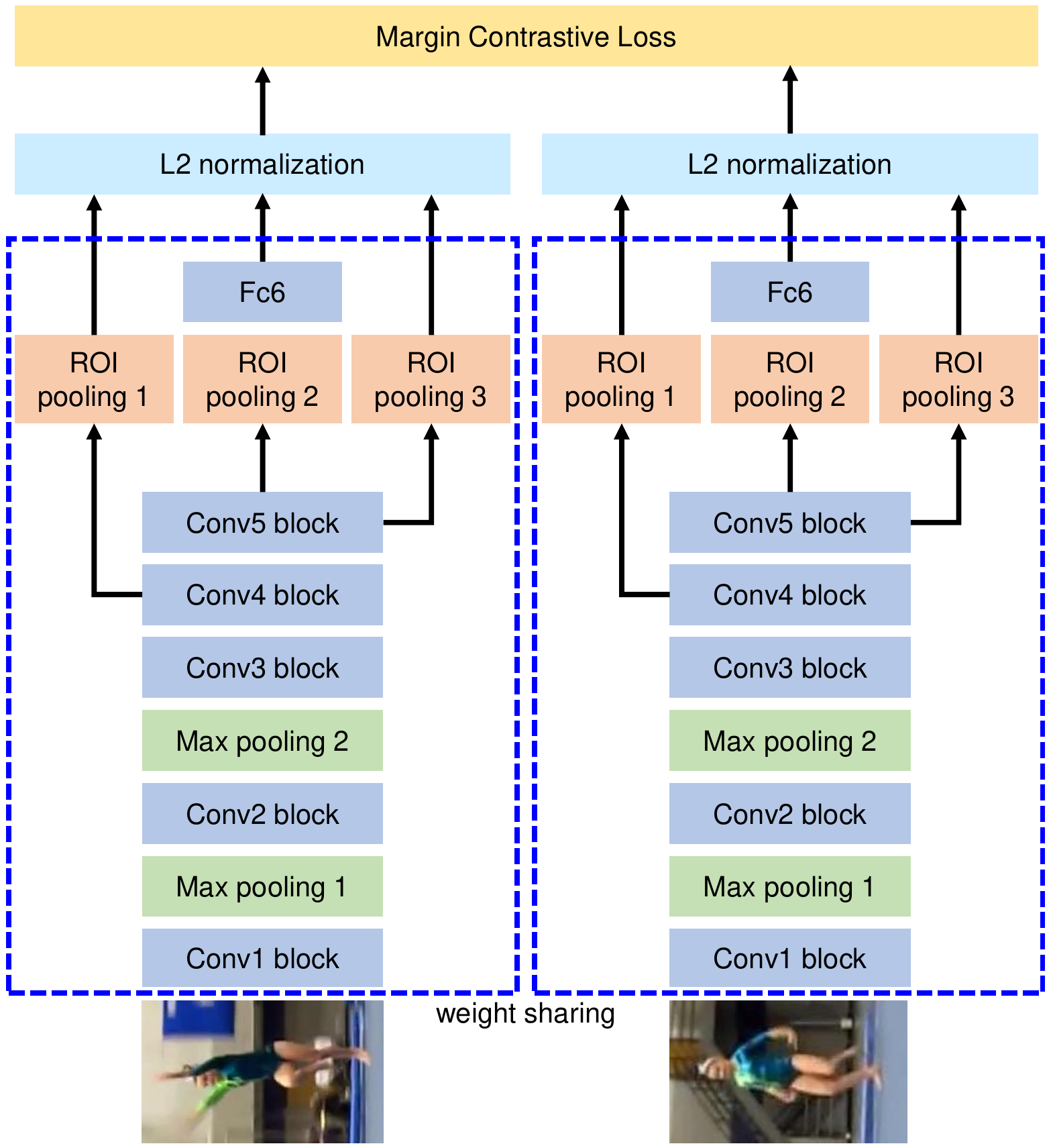}\\
\caption{Illustration of the architecture of the Siamese network for verifier.}
\label{siamese_net}
\end{figure}

\subsubsection{Verifying}

The goal of verifying is to measure the similarity between a given sample and the target object. Inspired by~\cite{chopra2005learning}, we use the Siamese network to develop verifier $\VF$ (similar to~\cite{tao2016siamese}) in PTAV, as depicted in Figure \ref{siamese_net}. The Siamese network contains two branches of CNNs, and processes two inputs separately. In this work, we borrow the architecture from VGGNet~\cite{simonyan2014very} for CNNs, but with an additional region of interest (RoI) pooling layer~\cite{girshick2015fast}. This is because, for detection, $\VF$ needs to process multiple regions in an image, from them the candidate most similar to the target object is selected as the final result. For efficiency, RoI pooling is used for simultaneously processing a set of regions.

In the Siamese network, the two CNN branches are connected with a single contrastive loss layer
\begin{equation}
\mathcal{L}(\x_{i}, \x_{j}, r_{ij})=\frac{1}{2}r_{ij}D^{2}+\frac{1}{2}(1-r_{ij})\mathrm{max}(0,\varepsilon-D^{2})  \label{eq10}
\end{equation}
where $D=\left \| \psi(\x_{i})-\psi(\x_{j}) \right \|_{2}$ is the Euclidean metric in which $\psi(\cdot)$ represents feature transformation via the Siamese network, $r_{ij}\in{\{0,1\}}$ indicates that whether $\x_{i}$ and $\x_{j}$ are the same object or not, and $\varepsilon$ represents the minimum distance margin. 

Once training is finished\footnote{In this work we adopt the same strategy as in~\cite{tao2016siamese} to train the verifier. We refer readers to~\cite{tao2016siamese} for detailed training process.}, one can use the learned verifying function $\nu$ to compute verification score for each tracking result $\x'$ via
\begin{equation}
\nu(\x_\mathrm{obj}, \x') = \psi(\x_\mathrm{obj})^{T}\psi(\x')   \label{eq11}
\end{equation}
where $\x_\mathrm{obj}$ represents a fixed target template in the first frame. This strategy, as used in our preliminary work~\cite{fan2017parallel}, may meet problems when the target object undergoes large appearance variations or deformations. As a result, using a fixed target template for verification may be unreliable for distant subsequent tracking results. 

To alleviate this issue, we propose to employ a dynamic target template set $\mathcal{S}$ for adaptive verification using $k$-means clustering. More specifically, $\mathcal{S}$ is comprised of two components $\mathcal{S}_f$ and $\mathcal{S}_d$. The $\mathcal{S}_f=\{\x_\mathrm{obj}\}$ contains only the target template $\x_\mathrm{obj}$ in the first frame and is fixed during tracking. The set $\mathcal{S}_d$ is initially empty. During tracking, it is dynamically updated by collecting tracking results with high verification scores, as described later.

\begin{figure}[!t]
\centering
\begin{tabular}{c}
\includegraphics[width=.62\linewidth]{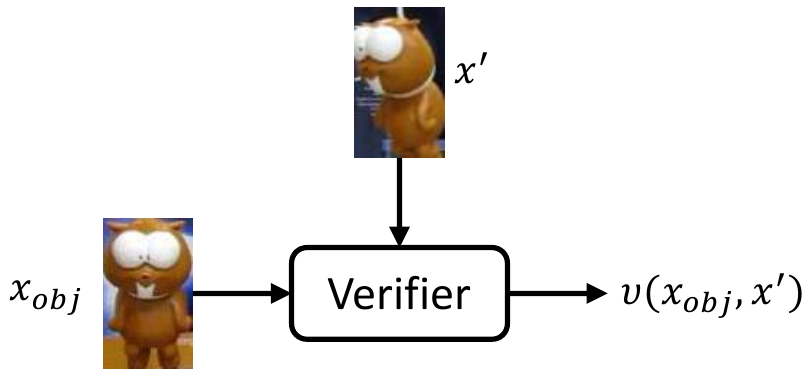}\\
(a) Verifying using a fixed target template in the first frame. \\
\includegraphics[width=\linewidth]{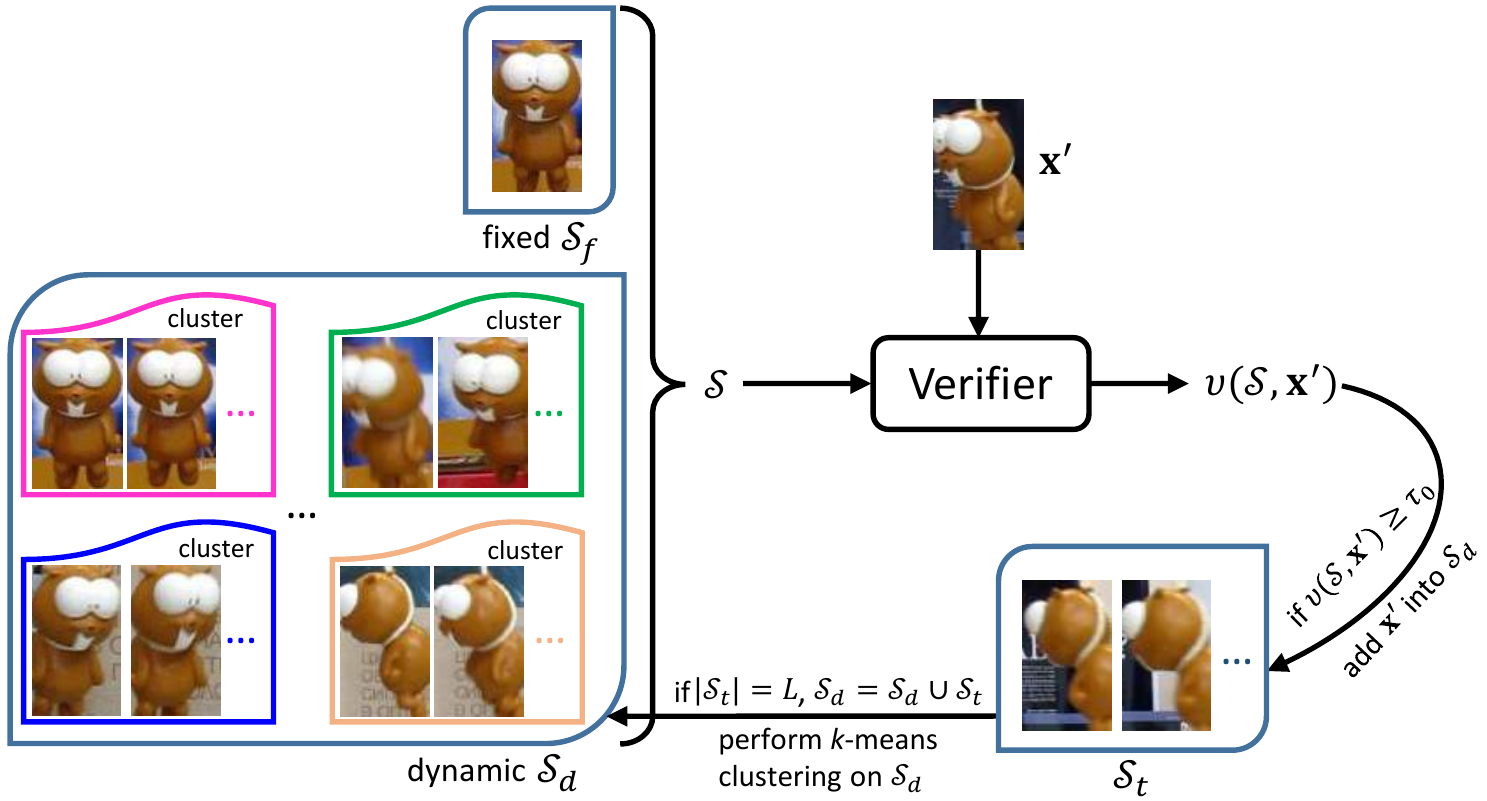}\\
(b) Verifying using the proposed dynamic target template set. \\
\end{tabular}
\caption{Illustration of different verifying strategies in~\cite{fan2017parallel} (see image (a)) and this paper (see image (b)).}
\label{target_pool}
\end{figure}

With the dynamic set $\mathcal{S}$, we can compute the verification score for each tracking result $\x'$ as follows
\begin{equation}
\nu(\mathcal{S},\x')=\omega_{o}\psi(\x_\mathrm{obj})^{T}\psi(\x')+\omega_{c}\sum\limits_{i=1}^{N_C}\sum\limits_{{\x_j\in{C_i}}}\psi(\x_{j})^{T}\psi(\x')  \label{eeq}
\end{equation}
where $\omega_{o}$ denotes the weight for  $\mathcal{S}_f$, $\omega_c$ represents the weight for each cluster $C_i$ obtained by performing $k$-means clustering on $\mathcal{S}_d$, and $N_C=|\mathcal{S}_d|/L$ is the number of clusters ($L$ is roughly a pre-defined size of each cluster, and $|\mathcal{S}_d|$ denotes the size of $\mathcal{S}_d$). The weights $\omega_{o}$ and $\omega_{c}$ are calculated as
\begin{eqnarray}
\omega_{o}  &=&  \frac{\mathrm{exp}(0.5)}{\mathrm{exp}(0.5)+{N_C}\times\mathrm{exp}(0.5/N_C)}  \label{eq13} \\
\omega_c  &=&  \frac{1}{N_C}(1-\omega_{o})
\label{eq14}
\end{eqnarray}

The set $\mathcal{S}_d$ is updated as follows. For each tracking result $\x'$, we use Equ. \ref{eeq} to calculate its verification score $\nu(\mathcal{S},\x')$. If $\nu(\mathcal{S},\x')$ is greater than a predefined threshold $\tau_0$, we treat $\x'$ as a reliable target template and add it into a temporal set $\mathcal{S}_t$. This process is repeated until the number of elements in $\mathcal{S}_t$ is equal to $L$. We then move all elements in $\mathcal{S}_t$ to $\mathcal{S}_d$ and thus leave $\mathcal{S}_t$ empty. If the number of elements in $\mathcal{S}_d$ is greater than $L\times{N_{C_\mathrm{max}}}$, where $N_{C_\mathrm{max}}$ denotes the maximum number of clusters, the oldest $L$ elements will be removed from $\mathcal{S}_d$. Afterwards, $k$-means clustering~\cite{kanungo2002efficient} is applied on $\mathcal{S}_d$ to obtain new clusters
\begin{equation}
\{C_i\}_{i=1}^{N_C}  = k \text{-} \mathrm{means} (\mathcal{S}_d, N_C)
\label{eeq1}
\end{equation}
Note that when performing $k$-means clustering, the elements in $\mathcal{S}_d$ are represented with HoG features~\cite{dalal2005histograms} for the sake of efficiency. After obtaining new clusters, we employ Equ. \ref{eq13} and \ref{eq14} to calculate weights $\omega_{o}$ and $\omega_{c}$. Figure \ref{target_pool} illustrates the process of adaptive verification.

With the dynamic target template set $\mathcal{S}$, $\mathcal{V}$ can make smarter decisions than when only a fixed template is used (i.e., Equ.~\ref{eq11}), and hence reduces the number of unnecessary verifications to speed up the entire system. Besides, now that verification is more precise, the verification-based detection (see Section \ref{obj_detection}) is improved as well. 

\subsubsection{Verification-based detection}
\label{obj_detection}
Given a tracking result from $\TRK$, we use Equ. \ref{eeq} to compute its verification score. If the verification score is lower than a predefined threshold $\tau_{1}$, $\VF$ will treat it as a tracking failure. In this case, $\VF$ needs to detect the target, again using the Siamese network. Unlike for verification, detection requires to verify multiple image patches from a local region\footnote{The local region is a square of size $\gamma(\mathrm{w}^{2}+\mathrm{h}^{2})^{\frac{1}{2}}$ centered at the location of the tracking result in this validation frame, where $\mathrm{w}$ and $\mathrm{h}$ denote the width and height of the tracking result, and $\gamma$ controls the scale and is dynamically adjusted based on detection result.} and finds the best one. Thanks to the RoI pooling layer, these candidates can be simultaneously processed in just one pass, resulting in significant reduction in computation. Let $\{\c_{i}\}_{i=1}^{N}$ denote the candidate set generated by sliding window, and the detection result $\widehat{\c}$ is determined by
\begin{equation}
\widehat{\c} = \argmax_{\c_{i}}{\;\nu(\mathcal{S}, \c_{i})}, \;\;\;\; i =1,2,\cdots,N
\end{equation}
where $\nu(\mathcal{S}, \c_{i})$ returns the verification score between the target template set $\mathcal{S}$ and candidate $\c_i$.

\begin{figure}[!t]
\centering
\includegraphics[width=\linewidth]{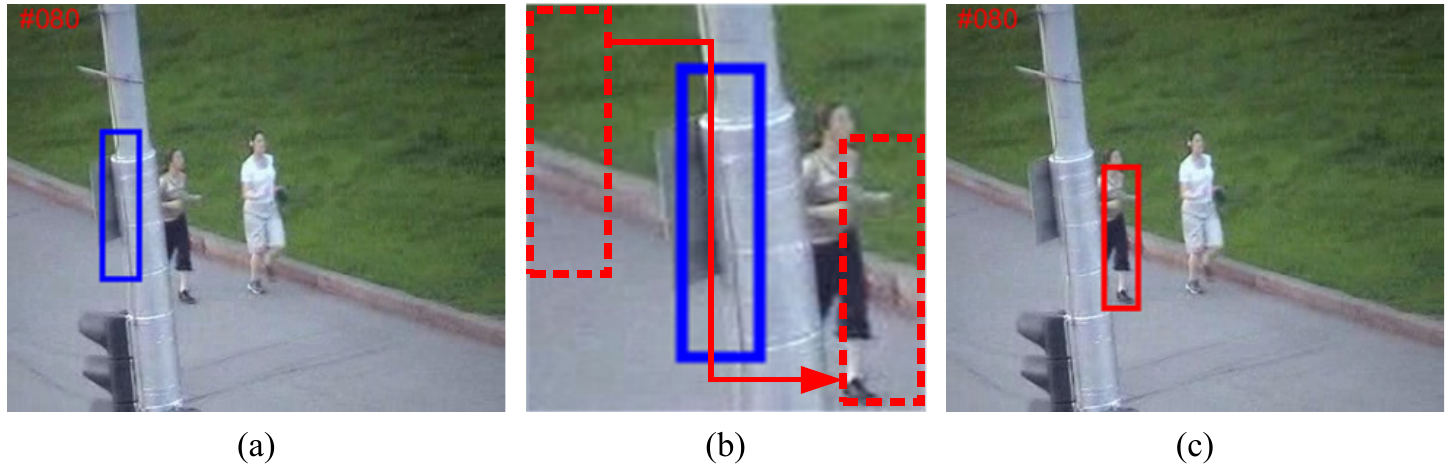}\\
\caption{Verification-based detection. When an unreliable tracking result is found (showing in \textcolor{blue}{blue} in (a)), the verifier $\VF$ searches/detects the target in a local region ( shown in (b)). The dashed \textcolor{red}{red} rectangles in (b) represent object candidates generated by sliding window. The \textcolor{red}{red} rectangle in (c) is the detection result.}
\label{detection}
\end{figure}

After obtaining the detection result $\widehat{\c}$, we determine whether or not to take it to be an alternative for tracking result according to its verification score. If $\nu(\mathcal{S},\widehat{\c})$ is less than a predefined threshold $\tau_{2}$, $\widehat{\c}$ is considered to be unreliable, and we do not replace tracking result with $\widehat{\c}$. Instead, we decrease the verifying interval $N_\mathrm{int}$ to 1, and enlarge the local searching region for target detection. Until detection result $\widehat{\c}$ passes verification (i.e., $\nu(\mathcal{S},\widehat{\c})\geq{\tau_{2}}$), we then restore $N_\mathrm{int}$ and the size of local searching region to initial settings. Figure \ref{detection} describes the detection process.

\subsection{Implementation Details}

Our PTAV is implemented in C++ and its verifier uses Caffe~\cite{jia2014caffe} on a single NVIDIA GTX TITAN Z GPU with 6GB memory. The merging factor $\alpha$ in Eq. (\ref{eq1}) is set to 0.3.
Other parameters for tracking remain the same as in~\cite{bertinetto2016staple}. The Siamese network for verification is initialized with the VGGNet~\cite{simonyan2014very} and trained based on the approach in~\cite{tao2016siamese}. The clustering interval $L$ is empirically set to 5 and the maximum number of clusters $N_{C_\mathrm{max}}$ to 10. The verification interval $N_{\mathrm{int}}$ is initially set to 10. The thresholds $\tau_0$, $\tau_1$ and $\tau_2$ are set to 0.6, 0.33 and 0.53, respectively. The parameter $\gamma$ is initialized to 1.5, and is adaptively adjusted based on the detection result. If the detection result with $\gamma=1.5$ is not reliable, the verifier will increase $\gamma$ for a larger searching region. Meanwhile, the verification interval $N_{\mathrm{int}}$ is decreased to 1. When the new detection result becomes faithful, $\gamma$ and $N_{\mathrm{int}}$ are then restored to 1.5 and 10. 

The source code of our implementation, as well as tracking results, are made publicly available at \url{http://www.dabi.temple.edu/~hbling/code/PTAV/ptav.htm}.

\section{Experiments}
\label{exp}
\subsection{Experiment on OTB2015}

\noindent {\bf Dataset and evaluation settings.} The OTB2015 benchmark~\cite{wu2015object} contains 100 fully annotated challenging video sequences. These sequences are labeled based on 11 attributes, including deformation (DEF), occlusion (OCC), scale variation (SV), illumination variations (IV), motion blur (MB), fast motion (FM), background clutter (BC), out-of-view (OV), low resolution (LR), in-plane rotation (IPR) and out-of-plane rotation (OPR).

Following the protocol in~\cite{wu2015object}, we use three metrics, \emph{distance precision rate} (DPR), \emph{overlap success rate} (OSR) and \emph{center location error} (CLE), to evaluate different tracking algorithms. DPR demonstrates the percentage of frames whose estimated average center location errors are within the given threshold distance (e.g., 20 pixels) to groundtruth. OSR shows the percentage of successful frames at the threshold ranging from 0 to 1, and can be defined as the overlap score more than a fixed value (e.g., 0.5), where the overlap ratio is defined as score=(area($R_{GT}\cap{R_{T}}$)$/$area($R_{GT}\cup{R_{T}}$)) with the groundtruth $R_{GT}$ and tracking result $R_{T}$. CLE represents the Euclid distance between centers of tracking result and groundtruth.

\subsubsection{Overall performance}

We evaluate PTAV on OTB2015~\cite{wu2015object} and compare it with twelve state-of-the-art trackers from three typical categories: (\lowercase\expandafter{\romannumeral1}) deep feature-based tracking algorithms, including SINT~\cite{tao2016siamese}, HCF~\cite{ma2015hierarchical}, SiamFC~\cite{bertinetto2016fully}, HDT~\cite{qi2016hedged} and CFNet~\cite{valmadre2017end}; (\lowercase\expandafter{\romannumeral2}) correlation filter based trackers, including fDSST~\cite{danelljan2016discriminative}, LCT~\cite{ma2015long}, KCF~\cite{henriques2015high} and Staple~\cite{bertinetto2016staple}; and (\lowercase\expandafter{\romannumeral3}) other representative tracking methods, including TLD~\cite{kalal2012tracking}, MEEM~\cite{zhang2014meem} and Struck~\cite{hare2016struck}. We also note that there are other state-of-the-art tracking algorithms such as MDNet~\cite{nam2016learning}, SANet~\cite{fan2016sanet} and C-COT~\cite{danelljan2016beyond} (see Figure \ref{fig:speed-accuracy}). However, the speeds of these trackers are around 1 frames per second (fps). Since this work is focused on real-time object tracking, we compare PTAV with trackers whose speeds are no less than 10 fps, except for SINT~\cite{tao2016siamese} since it can be viewed as the baseline for tracking by verification. In particular, for SINT~\cite{tao2016siamese}, we use its tracking results without optical flow because no optical flow part is provided from the released source code. Another baseline for PTAV is the Staple tracker~\cite{bertinetto2016staple}, which provides the (fast) tracking part of PTAV. It is worth noting that other tracking algorithms may also be used for the tracking part in PTAV.

\begin{figure}[!t]
    \centering
	\includegraphics[width=0.5\linewidth]{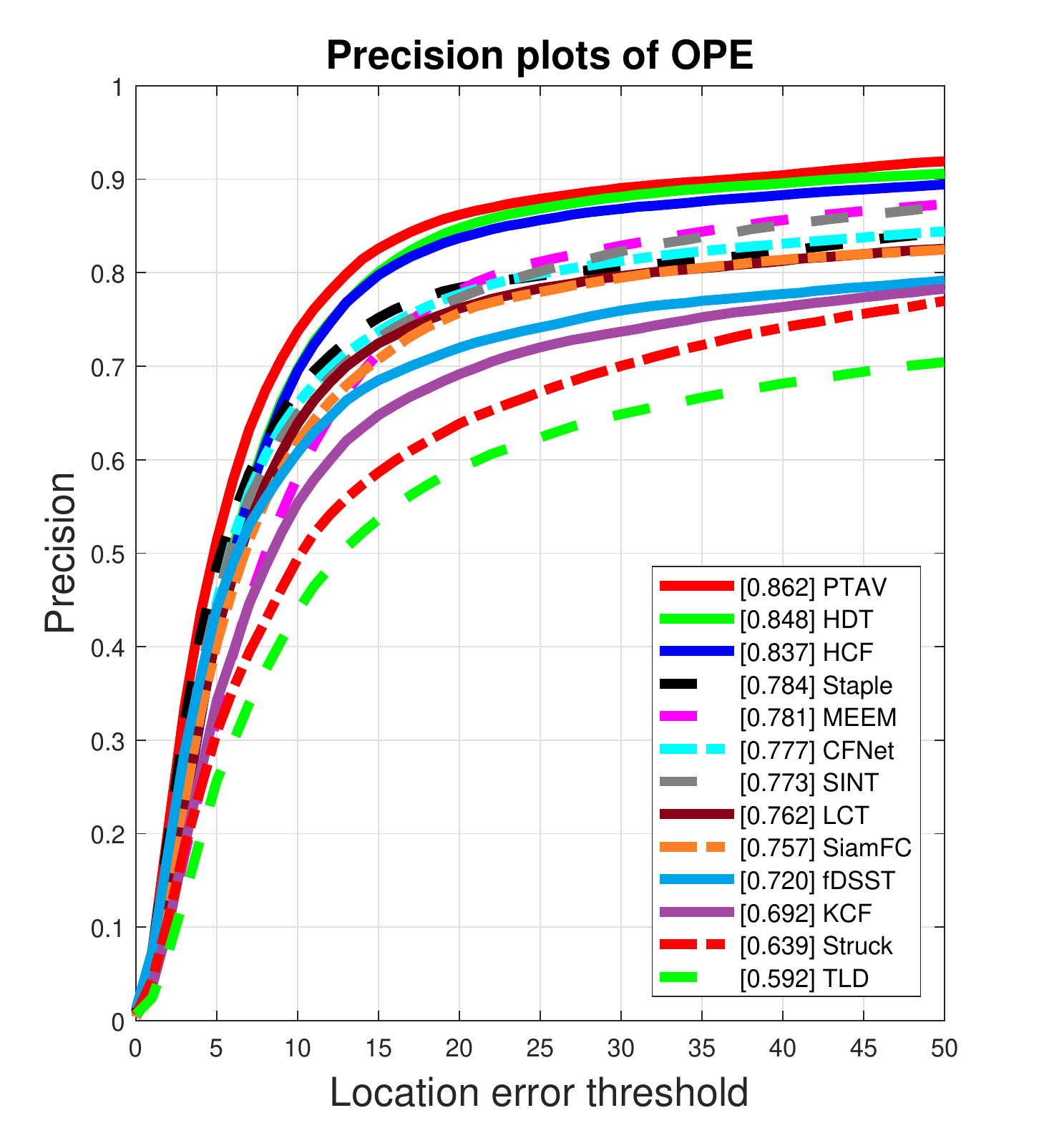}\includegraphics[width=0.5\linewidth]{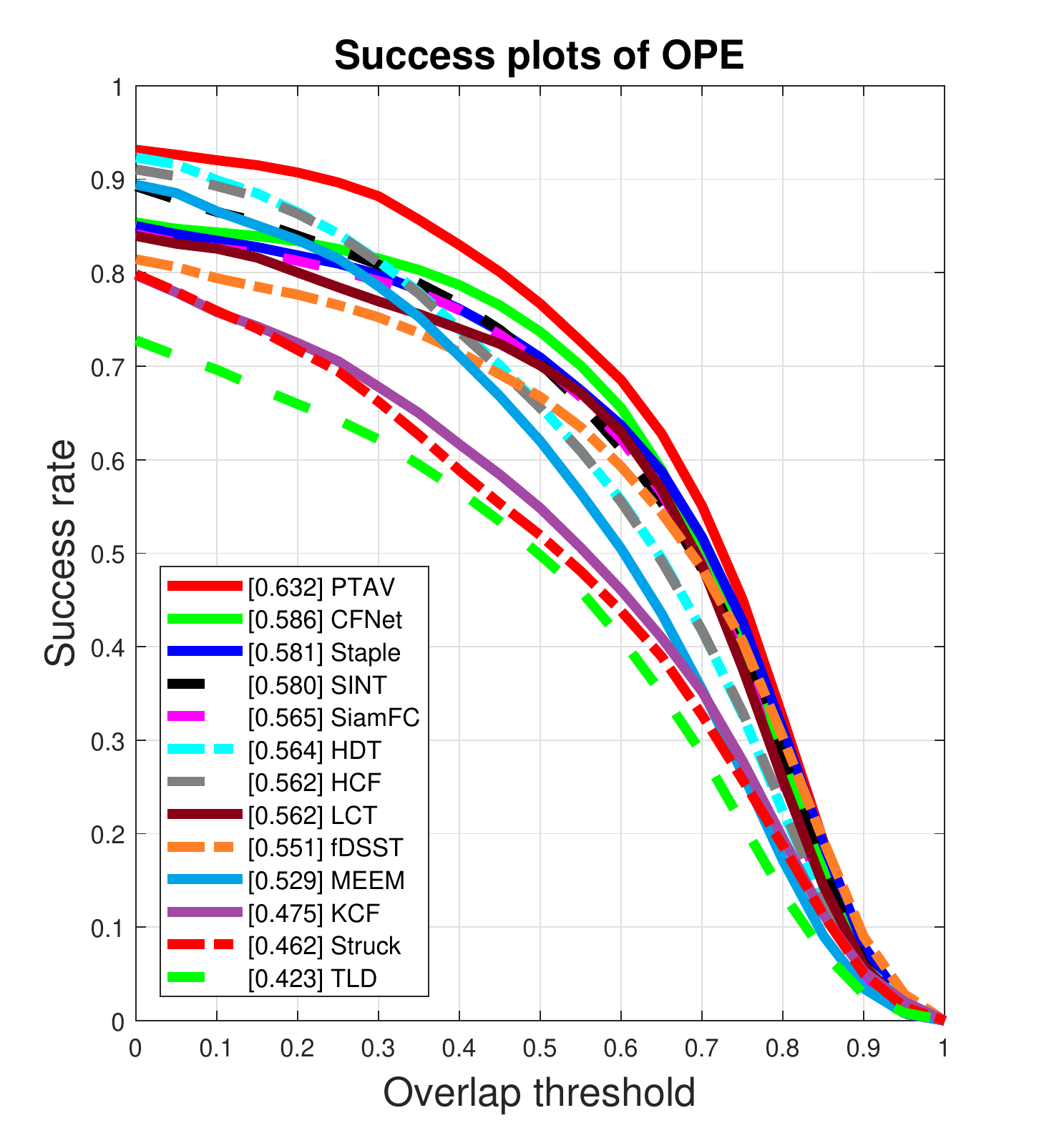}
	\caption{Comparison with pseudo real-time trackers on OTB2015\cite{wu2015object} using distance precision rate (DPR) and overlap success rate (OSR).}
	\label{comparison_OTB}
\end{figure}

\renewcommand\arraystretch{1.02}
\begin{table}[!t]
	\centering
	\caption{Comparisons with pseudo real-time tracking methods on OTB2015 \cite{wu2015object} in distance precision rate (DPR\%) at a threshold of 20 pixels, overlap success rate (OSR\%) at an overlap threshold of 0.5, center location error (CLE) in pixels and speed (fps). The best two results are highlighted in \textcolor{red}{red} and \textcolor{blue}{blue} fonts, respectively}
	\begin{tabular}{@{}C{0.5cm}|@{}R{1.8cm}C{1.1cm}@{}C{1.1cm}@{}C{1.1cm}@{}C{1.1cm}@{}}
		\hline
		\multicolumn{1}{r}{} & Algorithms & \multicolumn{1}{c}{DPR} & \multicolumn{1}{c}{OSR} & \multicolumn{1}{c}{CLE} & \multicolumn{1}{c}{Speed}\\
		\hline
		\hline
		\multicolumn{1}{r}{} & PTAV (Ours)    & \textcolor{red}{86.2}    & \textcolor{red}{77.9}    & \textcolor{red}{18.9}    & 27\\
		\hline
		\multirow{5}[0]{*}{\rotatebox{90}{(\lowercase\expandafter{\romannumeral1})}} & HCF~\cite{ma2015hierarchical}     & 83.7    & 65.6    & 22.8    & 10 \\
		& HDT~\cite{qi2016hedged}     & \textcolor{blue}{84.8}    & 64.9    & \textcolor{blue}{20.6}    & 10 \\
		& SINT~\cite{tao2016siamese}    & 77.3    & 70.3    & 26.3    & 2 \\
		& SiamFC~\cite{bertinetto2016fully}  & 75.7    & 70.9    & 37.1    & \textcolor{blue}{58} \\
		& CFNet~\cite{valmadre2017end}     & 77.7    & 73.2    & 35.2      & 43 \\
		\hline
		\multirow{5}[0]{*}{\rotatebox{90}{(\lowercase\expandafter{\romannumeral2})}}
		& Staple~\cite{bertinetto2016staple}  & 78.4    & 70.9    & 31.9    & 43 \\
		& LCT~\cite{ma2015long}     & 76.2    & 70.1    & 67.1    & 25 \\
		& fDSST~\cite{danelljan2016discriminative}   & 72.0    & 67.6    & 51.1    & 51 \\
		& KCF~\cite{henriques2015high}     & 69.2    & 54.8    & 45.0      & \textcolor{red}{243} \\
		\hline
		\multirow{4}[0]{*}{\rotatebox{90}{(\lowercase\expandafter{\romannumeral3})}} & MEEM~\cite{zhang2014meem}    & 78.1    & 62.2    & 27.7    & 21 \\
		& TLD~\cite{kalal2012tracking}    & 59.2   & 48.3    & 35.0    & 20 \\
		& Struck~\cite{hare2016struck}  & 63.9    & 51.6    & 47.1    & 10 \\
		\hline
	\end{tabular}%
	\label{OTB_table}%
\end{table}%

\begin{table*}[!htbp]
	\centering
	\caption{Average DPR (\%) in terms of individual attributes on OTB2015~\cite{wu2015object}. The best two results are highlighted in \textcolor{red}{red} and \textcolor{blue}{blue} fonts, respectively}
	\begin{tabular}{@{}C{1.3cm}@{}C{1cm}@{}C{1.15cm}@{}C{1.15cm}@{}C{1.3cm}@{}C{1.45cm}@{}C{1.35cm}@{}C{1.3cm}@{}C{1.45cm}@{}C{1.2cm}@{}C{1.35cm}@{}C{1.15cm}@{}C{1.35cm}@{}C{1.15cm}@{}}
		\hline
		Attributes & PTAV   & HDT~\cite{qi2016hedged}  & HCF~\cite{ma2015hierarchical}  & Staple~\cite{bertinetto2016staple} & MEEM~\cite{zhang2014meem} & CFNet~\cite{valmadre2017end}& SINT~\cite{tao2016siamese}& SiamFC~\cite{bertinetto2016fully}& LCT~\cite{ma2015long} & fDSST~\cite{danelljan2016discriminative}& KCF~\cite{henriques2015high}  & Struck~\cite{hare2016struck}& TLD~\cite{kalal2012tracking} \\
		\hline\hline
		IV    & \textcolor[rgb]{ 1,  0,  0}{84.7} & \textcolor[rgb]{ 0,  0,  1}{82.0} & 81.7  & 79.1  & 74.0  & 75.7  & 80.9  & 73.5  & 74.6  & 72.8  & 70.8  & 54.9  & 55.9 \\
		OPR   & \textcolor[rgb]{ 1,  0,  0}{83.5} & 80.8  & \textcolor[rgb]{ 0,  0,  1}{81.0} & 74.2  & 79.8  & 75.3  & 79.4  & 74.5  & 75.0  & 66.4  & 67.5  & 59.9  & 57.1 \\
		SV    & \textcolor[rgb]{ 1,  0,  0}{82.5} & \textcolor[rgb]{ 0,  0,  1}{81.1} & 80.2  & 73.1  & 74.0  & 74.8  & 74.2  & 74.3  & 68.6  & 66.9  & 63.9  & 60.4  & 56.4 \\
		OCC   & \textcolor[rgb]{ 1,  0,  0}{81.4} & \textcolor[rgb]{ 0,  0,  1}{77.4} & 76.7  & 72.6  & 74.1  & 71.3  & 73.1  & 69.6  & 68.2  & 62.6  & 62.2  & 53.3  & 52.4 \\
		DEF   & \textcolor[rgb]{ 0,  0,  1}{81.2} & \textcolor[rgb]{ 1,  0,  0}{82.1} & 79.1  & 74.8  & 75.4  & 66.9  & 75.0  & 67.6  & 68.9  & 59.9  & 61.7  & 52.7  & 48.4 \\
		MB    & \textcolor[rgb]{ 1,  0,  0}{80.5} & 79.4  & \textcolor[rgb]{ 0,  0,  1}{79.7} & 72.6  & 72.1  & 76.1  & 72.8  & 69.8  & 67.3  & 68.4  & 61.7  & 59.4  & 53.6 \\
		FM    & 78.1  & \textcolor[rgb]{ 1,  0,  0}{80.6} & \textcolor[rgb]{ 0,  0,  1}{79.7} & 70.3  & 73.4  & 74.1  & 72.5  & 73.0  & 67.5  & 69.3  & 62.8  & 62.0  & 54.8 \\
		IPR   & 82.8  & \textcolor[rgb]{ 0,  0,  1}{84.4} & \textcolor[rgb]{ 1,  0,  0}{85.4} & 77.0  & 79.3  & 80.3  & 81.1  & 74.8  & 78.2  & 72.5  & 69.3  & 63.4  & 60.3 \\
		OV    & \textcolor[rgb]{ 1,  0,  0}{79.0} & 66.3  & 67.7  & 66.1  & 68.3  & 65.0  & 72.5  & 67.8  & 59.2  & 57.7  & 49.8  & 49.1  & 45.2 \\
		BC    & \textcolor[rgb]{ 1,  0,  0}{87.6} & \textcolor[rgb]{ 0,  0,  1}{84.7} & \textcolor[rgb]{ 0,  0,  1}{84.7} & 77.0  & 75.1  & 73.7  & 75.1  & 69.4  & 74.0  & 78.4  & 71.6  & 57.3  & 46.1 \\
		LR    & \textcolor[rgb]{ 0,  0,  1}{84.0} & 76.6  & 78.7  & 60.9  & 60.5  & \textcolor[rgb]{ 1,  0,  0}{86.1} & 78.8  & 83.4  & 49.0  & 61.7  & 54.5  & 62.8  & 55.2 \\
		\hline\hline
		Overall & \textcolor[rgb]{ 1,  0,  0}{86.2} & \textcolor[rgb]{ 0,  0,  1}{84.8} & 83.7  & 78.4  & 78.1  & 77.7  & 77.3  & 75.7  & 76.2  & 72.0  & 69.2  & 63.9  & 59.2 \\
		\hline
	\end{tabular}%
	\label{OTB2015_attribute_DPR}%
\end{table*}%

\begin{table*}[!htbp]
	\centering
	\caption{Average OSR (\%) in terms of individual attributes on OTB2015~\cite{wu2015object}. The best two results are highlighted in \textcolor{red}{red} and \textcolor{blue}{blue} fonts, respectively.}
	\begin{tabular}{@{}C{1.3cm}@{}C{1cm}@{}C{1.15cm}@{}C{1.15cm}@{}C{1.3cm}@{}C{1.45cm}@{}C{1.35cm}@{}C{1.3cm}@{}C{1.45cm}@{}C{1.2cm}@{}C{1.35cm}@{}C{1.15cm}@{}C{1.35cm}@{}C{1.15cm}@{}}
		\hline
		Attributes & PTAV   & HDT~\cite{qi2016hedged}  & HCF~\cite{ma2015hierarchical}  & Staple~\cite{bertinetto2016staple} & MEEM~\cite{zhang2014meem} & CFNet~\cite{valmadre2017end}& SINT~\cite{tao2016siamese}& SiamFC~\cite{bertinetto2016fully}& LCT~\cite{ma2015long} & fDSST~\cite{danelljan2016discriminative}& KCF~\cite{henriques2015high}  & Struck~\cite{hare2016struck}& TLD~\cite{kalal2012tracking} \\
		\hline\hline
		IV    & \textcolor[rgb]{ 1,  0,  0}{64.2} & 53.5  & 54.0  & 59.8  & 51.7  & 57.4  & \textcolor[rgb]{ 0,  0,  1}{61.8} & 54.9  & 56.6  & 55.6  & 47.4  & 42.0  & 41.4 \\
		OPR   & \textcolor[rgb]{ 1,  0,  0}{60.4} & 53.6  & 53.7  & 53.8  & 52.8  & 55.3  & \textcolor[rgb]{ 0,  0,  1}{58.6} & 54.4  & 54.1  & 50.1  & 45.4  & 42.7  & 39.0 \\
		SV    & \textcolor[rgb]{ 1,  0,  0}{59.1} & 48.9  & 48.8  & 52.9  & 47.3  & 55.5  & \textcolor[rgb]{ 0,  0,  1}{55.8} & 55.5  & 49.2  & 51.0  & 39.9  & 40.7  & 38.8 \\
		OCC   & \textcolor[rgb]{ 1,  0,  0}{60.6} & 52.8  & 52.5  & 54.8  & 50.3  & 53.6  & \textcolor[rgb]{ 0,  0,  1}{55.8} & 52.3  & 50.7  & 47.8  & 43.8  & 39.3  & 36.3 \\
		DEF   & \textcolor[rgb]{ 1,  0,  0}{59.9} & 54.3  & 53.0  & 55.4  & 48.9  & 49.2  & \textcolor[rgb]{ 0,  0,  1}{55.5} & 49.0  & 49.9  & 46.1  & 43.6  & 38.3  & 34.1 \\
		MB    & \textcolor[rgb]{ 1,  0,  0}{61.2} & 56.3  & 57.3  & 55.8  & 54.3  & \textcolor[rgb]{ 0,  0,  1}{59.3} & 57.4  & 55.5  & 53.2  & 54.8  & 45.6  & 46.1  & 42.6 \\
		FM    & \textcolor[rgb]{ 1,  0,  0}{58.3} & 55.4  & 55.5  & 54.1  & 52.8  & \textcolor[rgb]{ 0,  0,  1}{57.0} & 55.7  & 56.4  & 52.7  & 55.4  & 45.5  & 46.1  & 41.8 \\
		IPR   & \textcolor[rgb]{ 1,  0,  0}{59.0} & 55.5  & 55.9  & 55.2  & 52.8  & \textcolor[rgb]{ 1,  0,  0}{59.0} & \textcolor[rgb]{ 0,  0,  1}{58.5} & 55.7  & 55.7  & 54.5  & 46.5  & 45.2  & 42.5 \\
		OV    & \textcolor[rgb]{ 1,  0,  0}{56.9} & 47.2  & 47.4  & 48.1  & 48.4  & 48.0  & \textcolor[rgb]{ 0,  0,  1}{55.9} & 51.1  & 45.2  & 45.7  & 39.3  & 37.8  & 33.5 \\
		BC    & \textcolor[rgb]{ 1,  0,  0}{64.1} & 58.0  & \textcolor[rgb]{ 0,  0,  1}{58.7} & 57.4  & 52.1  & 54.5  & 56.7  & 50.4  & 55.3  & 58.5  & 49.8  & 44.2  & 35.2 \\
		LR    & 54.6 & 42.0  & 42.4  & 41.1  & 35.5  & \textcolor[rgb]{ 1,  0,  0}{61.9} & 53.9  & \textcolor[rgb]{ 0,  0,  1}{60.4}  & 33.0  & 44.6  & 30.6  & 34.7  & 37.2 \\
		\hline\hline
		Overall & \textcolor[rgb]{ 1,  0,  0}{63.2} & 56.4  & 56.2  & 58.1  & 52.9  & \textcolor[rgb]{ 0,  0,  1}{58.6} & 58.0  & 56.5  & 56.2  & 55.1  & 47.5  & 46.2  & 42.3 \\
		\hline
	\end{tabular}%
	\label{OTB2015_attribute_OSR}%
\end{table*}%

We report the results in one-pass evaluation (OPE) using DPR and OSR as shown in Figure~\ref{comparison_OTB}. Overall, PTAV performs favorably against other tracking algorithms. In addition, we present quantitative comparison of DPR at 20 pixels, OSR at 0.5, center location error (CLE) in pixels and tracking speed (fps) in Table~\ref{OTB_table}. It demonstrates that PTAV outperforms other trackers in all three metrics. Among the trackers under comparison, HCF~\cite{ma2015hierarchical} utilizes deep hierarchical features to represent object appearance and obtains the DPR of 83.7\% and OSR of 65.6\%. Likewise, HDT~\cite{qi2016hedged} exploits all layers in VGGNet~\cite{simonyan2014very} for tracking and achieves the DPR of 84.8\% and OSR of 64.8\%. Compared to these two deep feature-based approaches, our tracker achieves better performance with DPR of 86.2\% and OSR of 77.9\%. Besides, owing to the adoption of parallel framework, PTAV (27 fps) is more than twice faster than the HCF~\cite{ma2015hierarchical} (10 fps) and HDT~\cite{qi2016hedged} (10 fps). Compared with SINT~\cite{tao2016siamese}, which uses similar Siamese network for tracking, PTAV improves DPR from 77.3\% to 86.2\% and OSR from 70.3\% to 77.6\%. In addition, PTAV runs at real-time while SINT~\cite{tao2016siamese} needs large improvement in speed. Compared with the baseline Staple~\cite{bertinetto2016staple}, PTAV achieves significant improvements on DPR (by 7.8\%) and OSR (by 7.0\%). Compared to representative MEEM~\cite{zhang2014meem} with DPR of 78.1\% and OSR of 62.2\%, PTAV obtains performance gains by DPR of 8.1\% and OSR of 15.7\%.

\begin{figure*}[!htb]
\centering
\begin{tabular}{@{\hspace{.0mm}}c@{\hspace{1.85mm}} @{\hspace{.0mm}}c@{\hspace{.0mm}}}
\includegraphics[width=2.1cm, height=1.575cm]{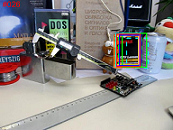} \includegraphics[width=2.1cm, height=1.575cm]{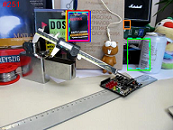} \includegraphics[width=2.1cm, height=1.575cm]{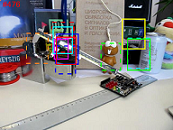} \includegraphics[width=2.1cm, height=1.575cm]{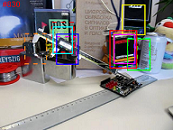} & \includegraphics[width=2.1cm, height=1.575cm]{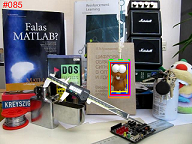} \includegraphics[width=2.1cm, height=1.575cm]{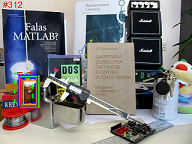} \includegraphics[width=2.1cm, height=1.575cm]{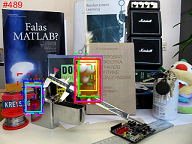} \includegraphics[width=2.1cm, height=1.575cm]{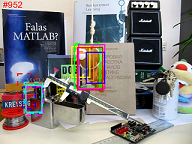} \\
\includegraphics[width=2.1cm, height=1.575cm]{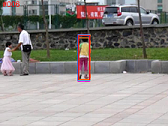} \includegraphics[width=2.1cm, height=1.575cm]{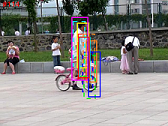} \includegraphics[width=2.1cm, height=1.575cm]{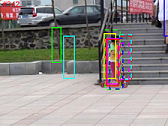} \includegraphics[width=2.1cm, height=1.575cm]{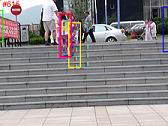} & \includegraphics[width=2.1cm, height=1.575cm]{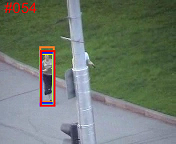} \includegraphics[width=2.1cm, height=1.575cm]{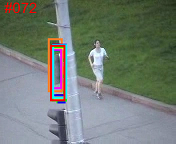} \includegraphics[width=2.1cm, height=1.575cm]{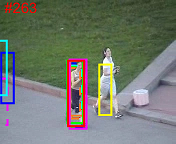} \includegraphics[width=2.1cm, height=1.575cm]{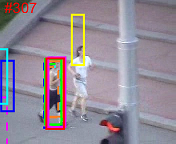} \\
\multicolumn{2}{c}{\small{(a) Sequences suffering from occlusions (from left to right and top to bottom: \emph{Box}, \emph{Lemming}, \emph{Girl2} and \emph{Jogging-1}). } }\\
\includegraphics[width=2.1cm, height=1.575cm]{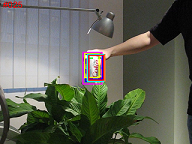} \includegraphics[width=2.1cm, height=1.575cm]{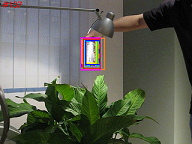} \includegraphics[width=2.1cm, height=1.575cm]{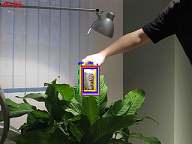} \includegraphics[width=2.1cm, height=1.575cm]{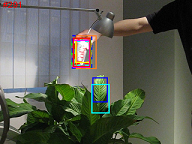} & \includegraphics[width=2.1cm, height=1.575cm]{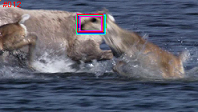} \includegraphics[width=2.1cm, height=1.575cm]{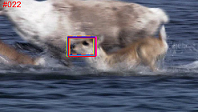} \includegraphics[width=2.1cm, height=1.575cm]{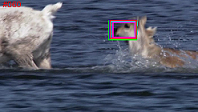} \includegraphics[width=2.1cm, height=1.575cm]{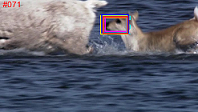} \\
\includegraphics[width=2.1cm, height=1.575cm]{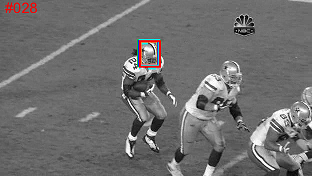} \includegraphics[width=2.1cm, height=1.575cm]{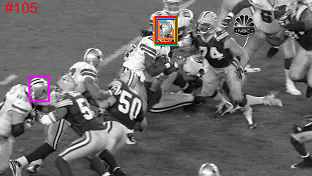} \includegraphics[width=2.1cm, height=1.575cm]{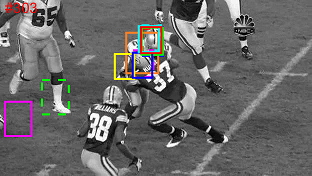} \includegraphics[width=2.1cm, height=1.575cm]{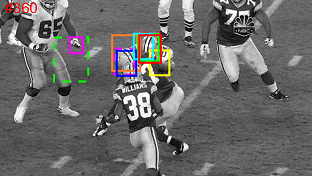} & \includegraphics[width=2.1cm, height=1.575cm]{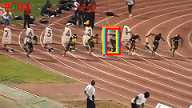} \includegraphics[width=2.1cm, height=1.575cm]{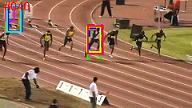} \includegraphics[width=2.1cm, height=1.575cm]{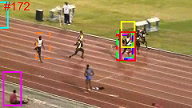} \includegraphics[width=2.1cm, height=1.575cm]{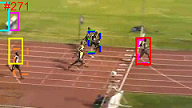} \\
\multicolumn{2}{c}{\small{(b) Sequences suffering from background clutters (from left to right and top to bottom: \emph{Coke}, \emph{Deer}, \emph{Football} and \emph{Bolt2}). } }\\
\includegraphics[width=2.1cm, height=1.575cm]{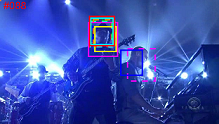} \includegraphics[width=2.1cm, height=1.575cm]{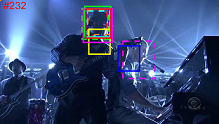} \includegraphics[width=2.1cm, height=1.575cm]{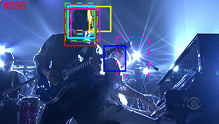} \includegraphics[width=2.1cm, height=1.575cm]{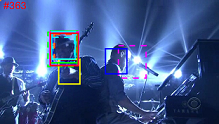} & \includegraphics[width=2.1cm, height=1.575cm]{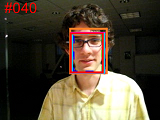} \includegraphics[width=2.1cm, height=1.575cm]{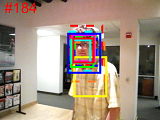} \includegraphics[width=2.1cm, height=1.575cm]{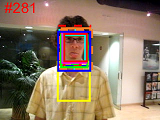} \includegraphics[width=2.1cm, height=1.575cm]{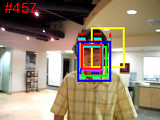} \\
\includegraphics[width=2.1cm, height=1.575cm]{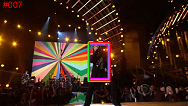} \includegraphics[width=2.1cm, height=1.575cm]{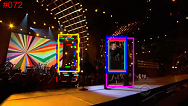} \includegraphics[width=2.1cm, height=1.575cm]{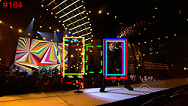} \includegraphics[width=2.1cm, height=1.575cm]{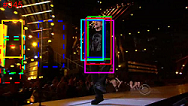} & \includegraphics[width=2.1cm, height=1.575cm]{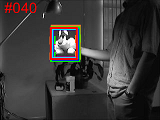} \includegraphics[width=2.1cm, height=1.575cm]{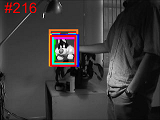} \includegraphics[width=2.1cm, height=1.575cm]{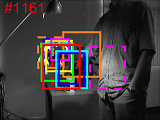} \includegraphics[width=2.1cm, height=1.575cm]{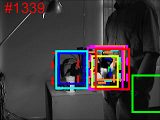} \\
\multicolumn{2}{c}{\small{(c) Sequences suffering from illumination variations (from left to right and top to bottom: \emph{Shaking}, \emph{David}, \emph{Singer2} and \emph{Sylvester}). } }\\
\includegraphics[width=2.1cm, height=1.575cm]{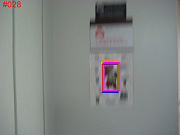} \includegraphics[width=2.1cm, height=1.575cm]{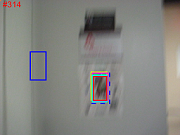} \includegraphics[width=2.1cm, height=1.575cm]{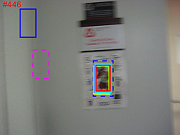} \includegraphics[width=2.1cm, height=1.575cm]{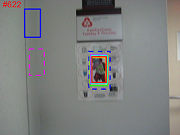} & \includegraphics[width=2.1cm, height=1.575cm]{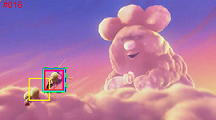} \includegraphics[width=2.1cm, height=1.575cm]{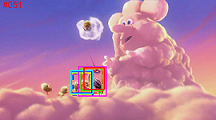} \includegraphics[width=2.1cm, height=1.575cm]{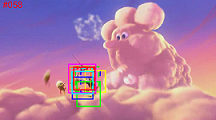} \includegraphics[width=2.1cm, height=1.575cm]{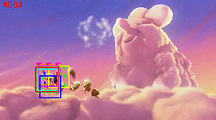} \\
\includegraphics[width=2.1cm, height=1.575cm]{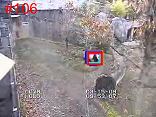} \includegraphics[width=2.1cm, height=1.575cm]{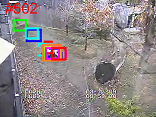} \includegraphics[width=2.1cm, height=1.575cm]{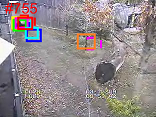} \includegraphics[width=2.1cm, height=1.575cm]{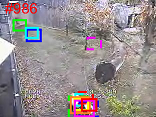} & \includegraphics[width=2.1cm, height=1.575cm]{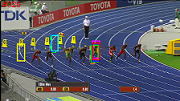} \includegraphics[width=2.1cm, height=1.575cm]{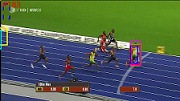} \includegraphics[width=2.1cm, height=1.575cm]{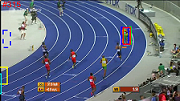} \includegraphics[width=2.1cm, height=1.575cm]{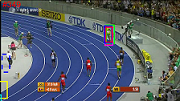} \\
\multicolumn{2}{c}{\small{(d) Sequences suffering from other challenges (from left to right and top to bottom: \emph{BlurOwl}, \emph{Bird2}, \emph{Panda} and \emph{Bolt}). } }\\
\multicolumn{2}{c}{\includegraphics[width=17cm]{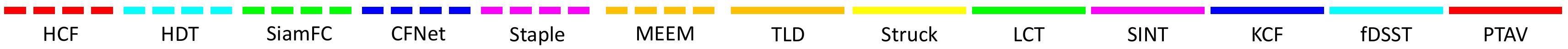}}\\
\end{tabular}
\caption{Qualitative evaluation of the proposed algorithm and other twelve state-of-the-art trackers on sixteen challenging sequences.}
\label{qua_res}
\end{figure*}

\subsubsection{Attribute-based evaluation}

We further analyze the performance of PTAV under the eleven different attributes on OTB2015. In Tables~\ref{OTB2015_attribute_DPR} and~\ref{OTB2015_attribute_OSR}, we summarize the evaluation results in terms of DPR and OSR.

For DPR, PTAV achieves the best results under 7 out of 11 attributes including IV (84.7\%), OPR (83.5\%), SV (82.5\%), OCC (81.4\%), MB (80.5\%), OV (79.0\%) and BC (87.6\%). For sequences with deformation, HDT~\cite{qi2016hedged} performs the best with average DPR of 82.1\% owing to the use of richer deep features. Our tracker uses Staple~\cite{bertinetto2016staple} as the tracking part, which leverages color information to handle object deformation. Accompanied by an accurate verifier, PTAV achieves competitive performance and ranks the second in the case of deformation with average DPR of 81.2\%. Furthermore, compared to the baseline Staple~\cite{bertinetto2016staple}, we obtain large gain on the average DPR by 6.4\% under deformation. For low resolution sequences, CFNet~\cite{valmadre2017end} obtains the best result with average DPR of 86.1\% by taking advantage of deep features. PTAV ranks the second with competitive average DPR of 84.0\%. 

For sequences with fast motion and in-plane rotation, the two deep feature-based trackers HDT~\cite{qi2016hedged} and HCF~\cite{ma2015hierarchical} perform better than ours, because in these two situations deep features are more efficacious than hand-crafted features to represent object appearance. In PTAV, we utilize simple HoG~\cite{dalal2005histograms} features and RGB histograms to model object appearance in tracking, which are sensitive to in-plane rotation and fast motion (we can see that from the performance of Staple~\cite{bertinetto2016staple}). Nevertheless, with the help of useful feedbacks from a robust and accurate verifier, PTAV still achieves competitive performance with average DPRs of 78.1\% and 82.8\% under these two challenges, respectively.

For OSR, on the other hand, PTAV achieves the best results under 10 of 11 attributes including IV (64.2\%), OPR (60.4\%), SV (59.1\%), OCC (60.6\%), DEF (59.9\%), MB (61.2\%), FM (58.3\%), IPR (59.0\%), OV (56.9\%) and BC (64.1\%). Low resolution (LR) is the only attribute for which PTAV does not rank the best, while CFNet~\cite{valmadre2017end} and SiamFC~\cite{bertinetto2016fully} obtain better results than PTAV. Specifically, PTAV achieves an OSR of 54.6\%, higher than all other trackers including its two baselines. 

\subsubsection{Qualitative evaluation}

To further analyze and demonstrate the performance of PTAV, we conduct rich qualitative evaluation described as following.

\vspace{0.5em}
\noindent {\bf Occlusions.} Figure \ref{qua_res}(a) demonstrates the sampled tracking results on sequences \emph{Box}, \emph{Lemming}, \emph{Girl2} and \emph{Jogging-1}, all involving heavy target occlusions. In \emph{Box}, the target undergoes not only occlusions but also background clutters (e.g., $\#$476). In \emph{Lemming}, the tracking target suffers from occlusions and motion blur (e.g., $\#$312 and $\#$489). In \emph{Girl2}, the target is fully occluded by the background (e.g., $\#$114). In sequence \emph{Jogging-1}, the tracking target is heavily occluded with significant deformation (e.g., $\#$72). From Figure \ref{qua_res}(a) we can see that PTAV handles well the occlusion in these sequences. Though the tracking part may lose the target temporally because of occlusions, it can quickly be corrected by the verifier and resumes tracking. Compared to deep feature based trackers (HCF~\cite{ma2015hierarchical}, HDT~\cite{qi2016hedged} and CFNet~\cite{valmadre2017end}), which lose the tracking target when occlusions happen (e.g., $\#$342 in \emph{Girl2} and $\#$489 in \emph{Lemming}), PTAV performs more robustly. Since correlation filter is sensitive to occlusion and no re-detection module is adopted, KCF~\cite{henriques2015high}, fDSST~\cite{danelljan2016discriminative} and Staple~\cite{bertinetto2016staple} lose the tracking target in all sequences. LCT~\cite{ma2015long} applies an additional detector to re-localize tracking target when it recovers from occlusion. Nevertheless, it still drifts to the background when occlusions are accompanied with background clutters (e.g., $\#$476 in \emph{Box}). MEEM~\cite{zhang2014meem} stores multiple memories of target during tracking, and re-detects the target using old memories when it recovers from occlusion (e.g., \emph{Lemming}, \emph{Girl2} and \emph{Jogging-1}). However, it meets problems in presence of background clutters (e.g., $\#$476 and $\#$830 in \emph{Box}). SiamFC~\cite{bertinetto2016fully} and SINT~\cite{tao2016siamese} deal with occlusion by searching in a local region when the target recovers from occlusions. In addition, no model update is adopted in SiamFC~\cite{bertinetto2016fully} and SINT~\cite{tao2016siamese}, avoiding introducing background into the tracking model. Other approaches such as TLD~\cite{kalal2012tracking} and Struck~\cite{hare2016struck} does not deal well with occlusion and drift to background (e.g., $\#$342 in \emph{Girl2}).

\vspace{0.5em}
\noindent {\bf Background clutters.} Background clutters are prone to result in tracking drift and even failures because the tracker may mix the tracking target with cluttering background. Figure \ref{qua_res}(b) displays the sampled experimental results on sequences \emph{Coke}, \emph{Deer}, \emph{Football} and \emph{Bolt2}. In \emph{Coke}, the object appearance is visually similar to the background, and occlusion and illumination variation occur as well. In \emph{Deer}, \emph{Football} and \emph{Bolt2}, there exist similar distracters involved with other challenges including occlusions (e.g., \emph{Football}), deformations (e.g., \emph{Bolt2}) and motion blur (e.g., \emph{Deer}). From Figure \ref{qua_res}(b), we can observe that SiamFC~\cite{bertinetto2016fully}, SINT~\cite{tao2016siamese}, KCF~\cite{henriques2015high}, Struck~\cite{hare2016struck} and Staple~\cite{bertinetto2016staple} drift to background in \emph{Football} (e.g., $\#$360). CFNet~\cite{valmadre2017end} and fDSST~\cite{danelljan2016discriminative} localize well the target in \emph{Football}, but still fail in \emph{Coke} (e.g., $\#$291) where background clutters are accompanied with occlusions and illumination change. MEEM~\cite{zhang2014meem} handles nicely occlusion and motion blur, but fails in presence of deformations in \emph{Bolt2} (e.g., $\#$40). LCT~\cite{ma2015long} and TLD~\cite{kalal2012tracking} are robust against background clutters since they utilize an extra object detector to seek for the target after it moves away from the cluttering region. However, they lose the tracking target when heavy deformation happens in \emph{Bolt2} (e.g., $\#$271). HCF~\cite{ma2015hierarchical} and HDT~\cite{qi2016hedged} perform robustly in these sequences because of the powerful deep features. Likewise, our tracker is able to deal with these situations owing to the verifier. Besides, the cooperation between tracking and verifying allows PTAV to run in real-time.

\vspace{0.5em}
\noindent {\bf Illumination variations.} Illumination variation often causes drift problem. Figure \ref{qua_res}(c) shows sampled results of sequences \emph{Shaking}, \emph{David}, \emph{Singer2} and \emph{Sylvester}. In \emph{Shaking} and \emph{Sylvester}, the target suffers from not only illumination variations but also background clutters and rotations. In \emph{David} and \emph{Singer2}, illumination variations are accompanied by rotations and scale changes. We can see from Figure \ref{qua_res}(c) that deep feature-based trackers SiamFC~\cite{bertinetto2016fully}, CFNet~\cite{valmadre2017end}, HCF~\cite{ma2015hierarchical} and HDT~\cite{qi2016hedged} lose the target in \emph{Singer2} (e.g., $\#$345) where background clutters happen. MEEM~\cite{zhang2014meem} can handle rotations but still fails in \emph{Singer2} (e.g., $\#$345). Staple~\cite{bertinetto2016staple} and KCF~\cite{henriques2015high} are sensitive to rotation and fail in \emph{Shaking} (e.g., $\#$363) and \emph{Sylvester} (e.g., $\#$1339). LCT works well in \emph{Shaking}, \emph{David} and \emph{Singer2}, yet have problems when heavy rotation exists in \emph{Sylvester} (e.g., $\#$1339). Our tracker performs well on these sequences. Though its tracking part may drift to background due to rotation (e.g., $\#$1161 in \emph{Sylvester}), this situation is found and immediately corrected by its verifier (e.g., $\#$1339 in \emph{Sylvester}).

\vspace{0.5em}
\noindent {\bf Other challenges.} Figure \ref{qua_res}(d) demonstrates results of sequences \emph{BlurOwl}, \emph{Bird2}, \emph{Panda} and \emph{Bolt2}, which contain other challenges including fast motion, motion blur, rotation, scale change, deformation and so forth. In \emph{BlurOwl}, the camera moves quickly, causing serious motion blur (e.g., $\#$622). KCF~\cite{henriques2015high} and Staple~\cite{bertinetto2016staple} lose the target, while PTAV well localizes the tracking target thanks to its verifier which corrects tracker. In \emph{Bolt2}, TLD~\cite{kalal2012tracking}, Struck~\cite{hare2016struck}, CFNet~\cite{valmadre2017end} and fDSST~\cite{danelljan2016discriminative} drift to background because of deformations. On \emph{Panda}, LCT~\cite{ma2015long}, fDSST~\cite{danelljan2016discriminative}, Staple~\cite{bertinetto2016staple} and KCF~\cite{henriques2015high} lose the tracking target owing to scale changes and rotations. By contrast, MEEM~\cite{zhang2014meem}, HCF~\cite{ma2015hierarchical}, HDT~\cite{qi2016hedged}, SiamFC~\cite{bertinetto2016fully}, SINT~\cite{tao2016siamese} and our PTAV performs favorably because of powerful feature representation.

\subsection{Experiment on TC128}

\begin{figure}[!t]
	\centering
	\includegraphics[width=0.5\linewidth]{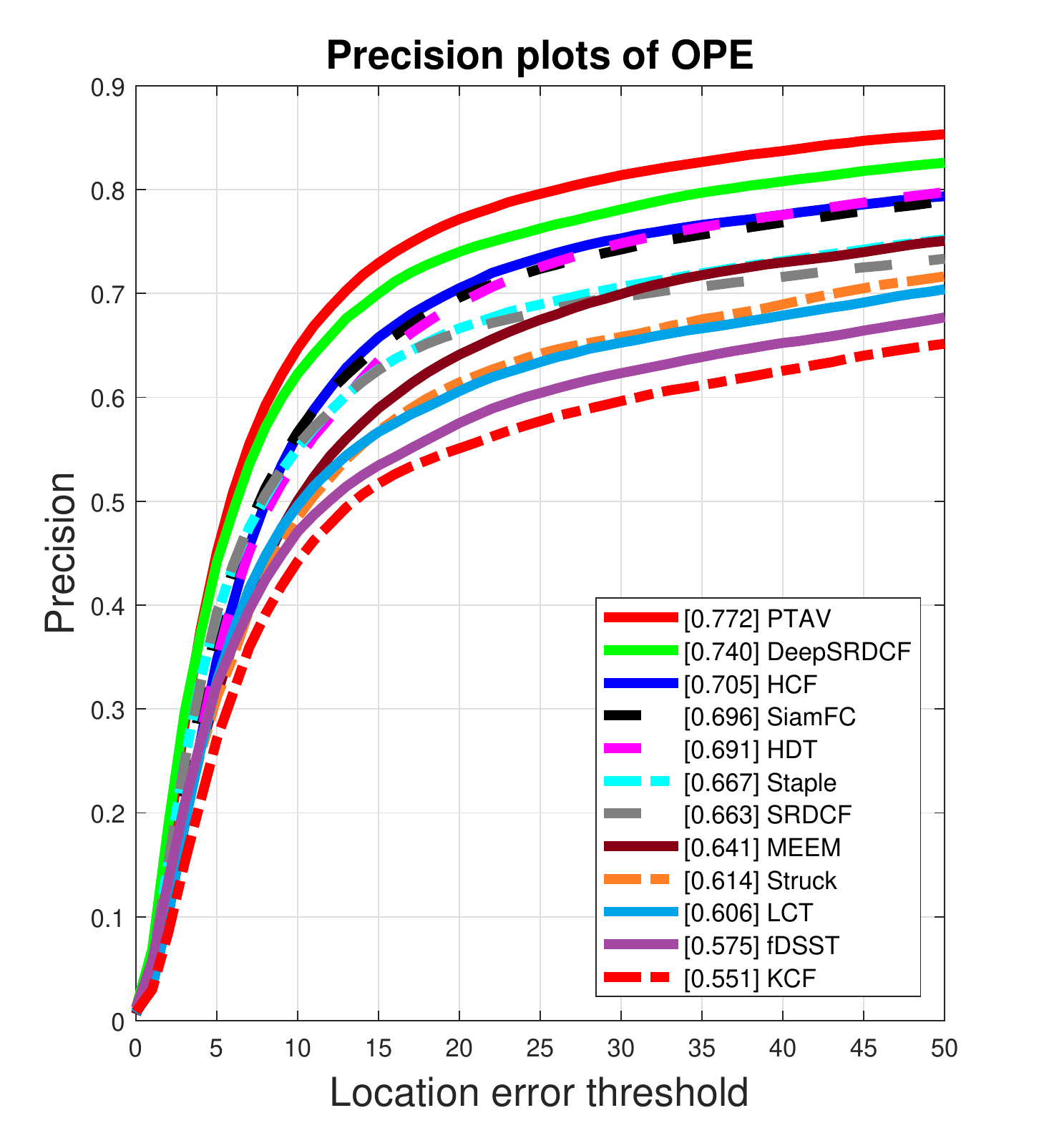}\includegraphics[width=0.5\linewidth]{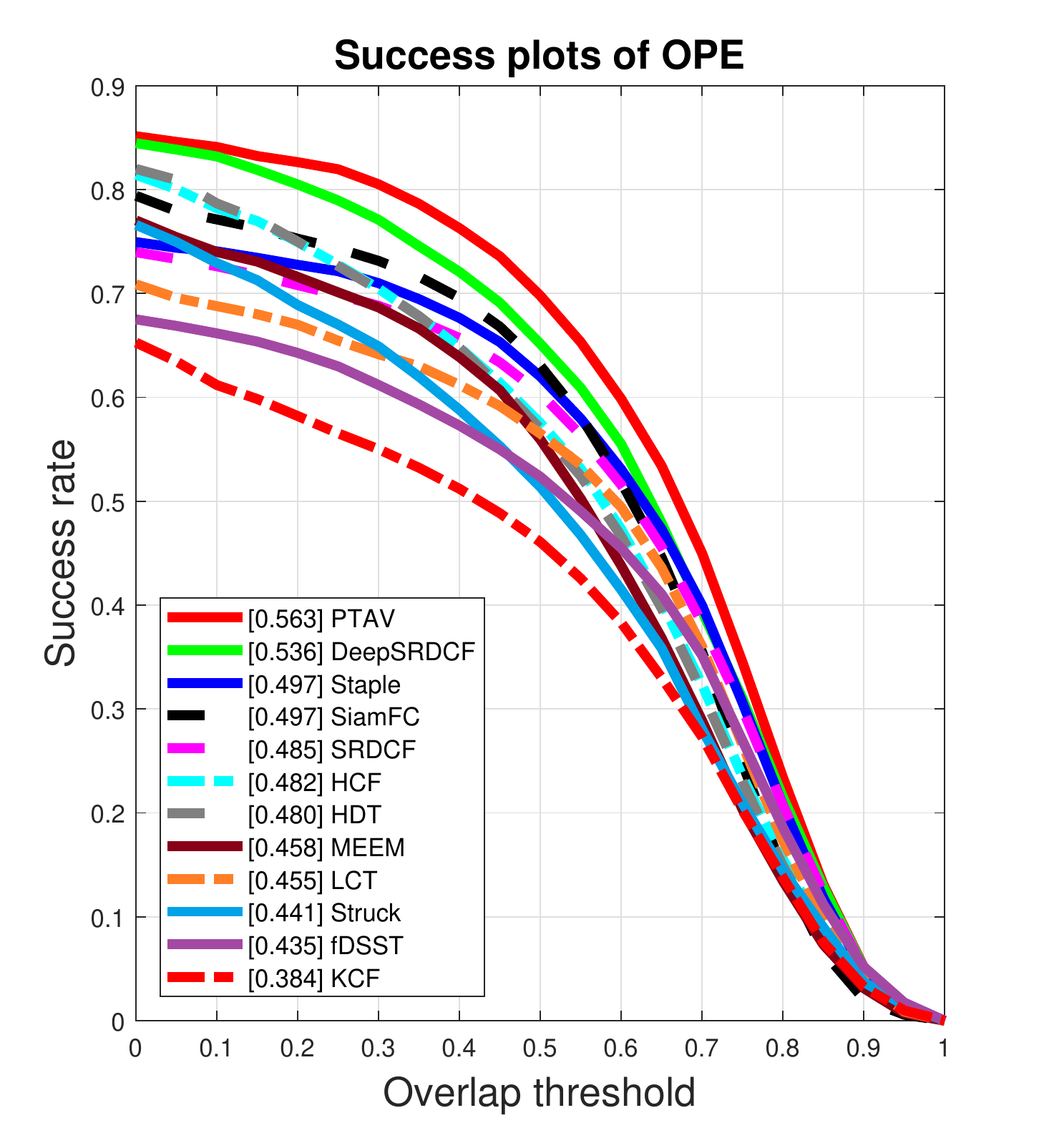}
	\caption{Comparison with eleven state-of-the-art trackers on TC128~\cite{liang2015encoding} using distance precision rate and overlap success rate.}
	\label{comparison_TC}
\end{figure}

The TC128 benchmark~\cite{liang2015encoding} is comprised of 128 fully annotated challenging color sequences. On TC128~\cite{liang2015encoding}, PTAV runs at 24 \emph{fps} and is with eleven state-of-the-art trackers including MEEM~\cite{zhang2014meem}, HCF~\cite{ma2015hierarchical}, HDT~\cite{qi2016hedged}, Staple~\cite{bertinetto2016staple}, SiamFC~\cite{bertinetto2016fully}, SRDCF~\cite{danelljan2015learning}, DeepSRDCF~\cite{danelljan2015convolutional}, fDSST~\cite{danelljan2016discriminative}, KCF~\cite{henriques2015high}, LCT~\cite{ma2015long} and Struck~\cite{hare2016struck}. Following the protocol in~\cite{liang2015encoding}, we report evaluation results in OSR and DPR as shown in Figure \ref{comparison_TC}.

\begin{figure}[!t]
	\centering
	\includegraphics[width=\linewidth]{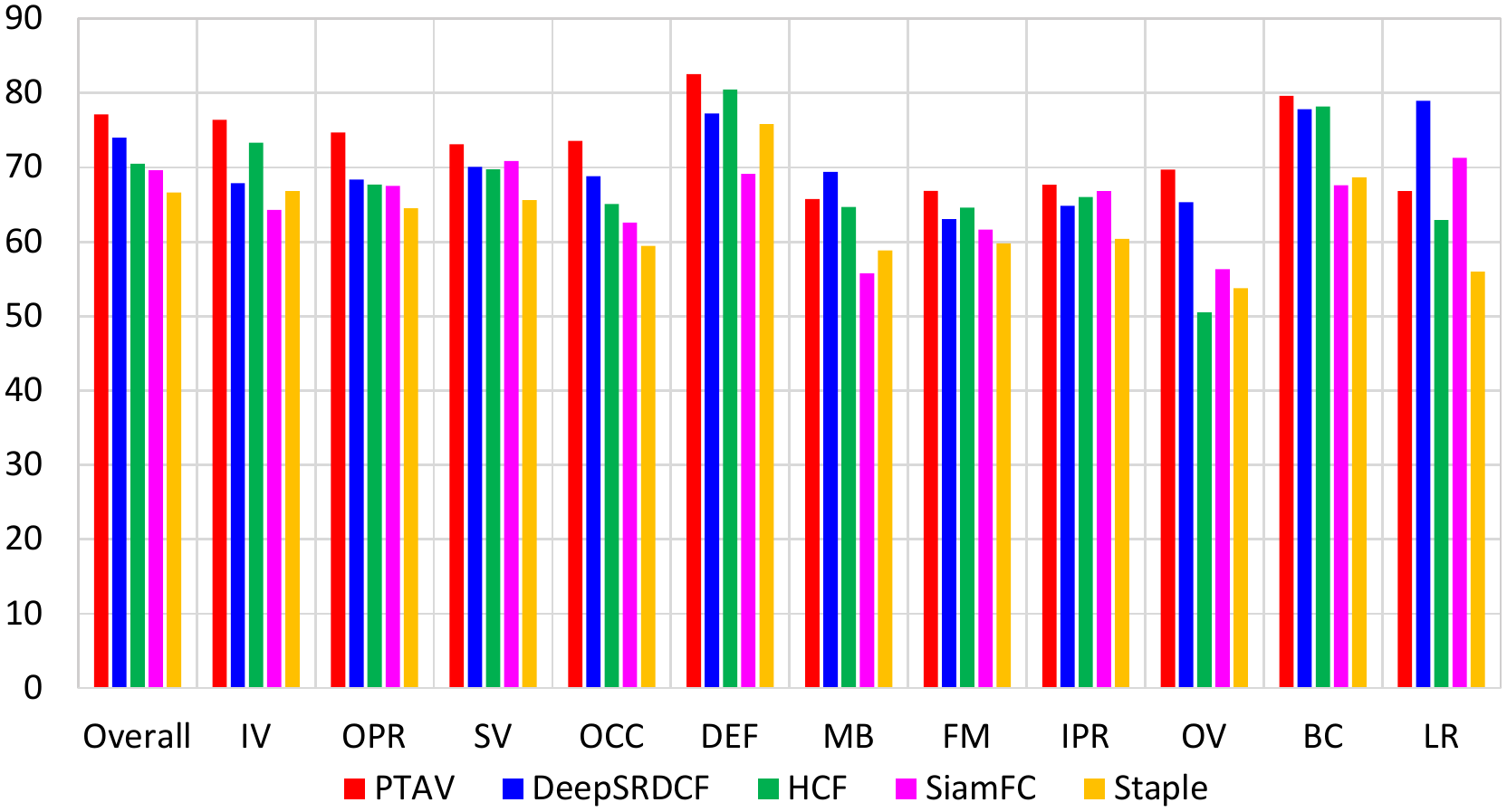}
	\caption{Average DPR (\%) in term of individual attributes on TC128~\cite{liang2015encoding}.}
	\label{TC_att}
\end{figure}

Among the eleven compared trackers, DeepSRDCF~\cite{danelljan2015convolutional} extends SRDCF~\cite{danelljan2015learning} by replacing hand-crafted features with convolutional features and obtains the best performance with DPR of 74.0\% and OSR of 53.6\%. By contrast, PTAV improves the state-of-the-art methods on DPR to 77.2\% and OSR to 56.3\%, obtaining the gains of 3.2\% and 2.7\%, respectively. In comparison with SiamFC~\cite{bertinetto2016fully} with DPR of 69.6\% and OSR of 49.7\%, PTAV achieves improvements of 7.2\% and 6.6\% on DPR and OSR, respectively. Compared with the other baseline, Staple~\cite{bertinetto2016staple}, which obtains a DPR of 66.7\% and an OSR of 49.7\%, PTAV achieves significant improvements as well, showing clearly the benefits of introducing a verifier.
For more detailed analysis, we show the average DPR for five trackers on different attributes in Figure \ref{TC_att}. PTAV can well handle various challenging factors and outperform the other four trackers in nine out of eleven attributes.

\subsection{Experiment on UAV20L}

The recently proposed UAV20L dataset~\cite{mueller2016benchmark} consists of 20 fully annotated sequences, with length ranging from 1,717 to 5,527 frames. These videos are challenging because the tracking target suffers from appearance changes caused by various factors. The proposed PTAF tracker runs at 30 \emph{fps}, and compared with ten state-of-the-art trackers including SiamFC~\cite{bertinetto2016fully}, MUSTer~\cite{hong2015multi}, SRDCF~\cite{danelljan2015learning}, HCF~ \cite{ma2015hierarchical}, MEEM~\cite{zhang2014meem}, SAMF~\cite{li2014scale}, Struck~\cite{hare2016struck}, fDSST~ \cite{danelljan2016discriminative}, LCT~\cite{ma2015long} and KCF~\cite{henriques2015high}.

\begin{figure}[!t]
	\centering
	\includegraphics[width=0.5\linewidth]{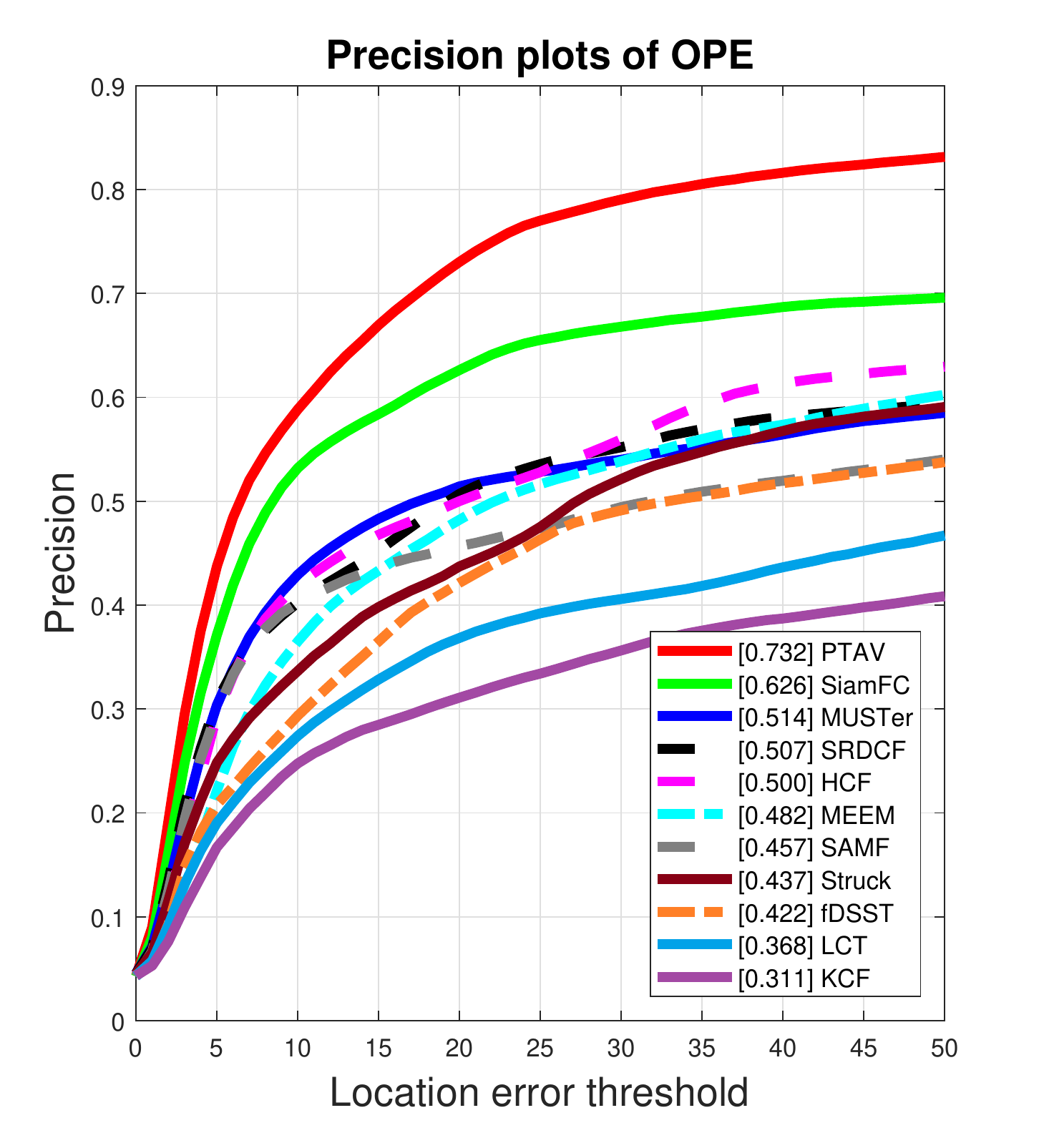}\includegraphics[width=0.5\linewidth]{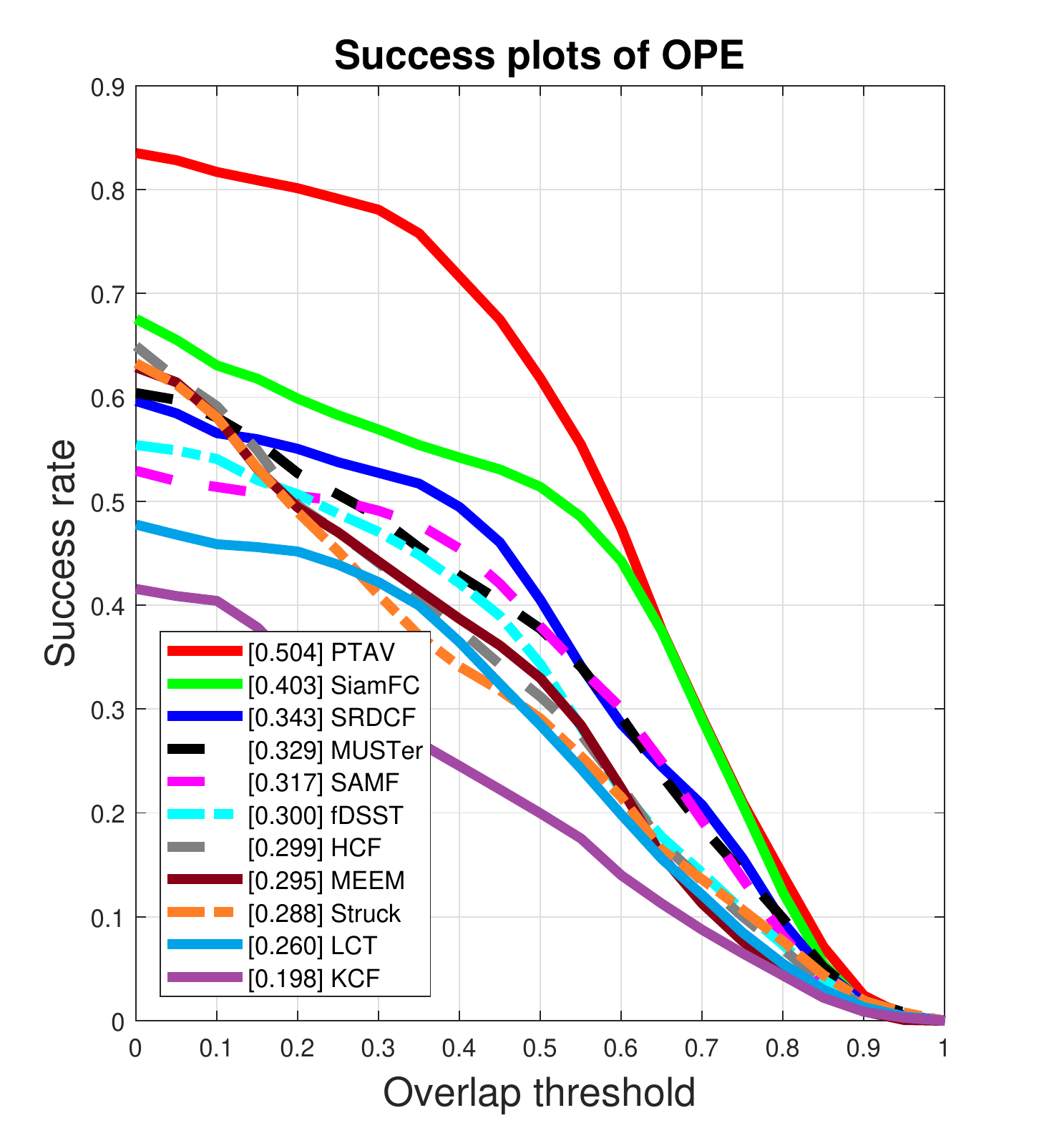}
	\caption{Comparison with ten state-of-the-art trackers on UAV20L~\cite{mueller2016benchmark} using distance precision rate and overlap success rate.}
	\label{comparison_UAL}
\end{figure}

\begin{figure}[!t]
	\centering
	\includegraphics[width=\linewidth]{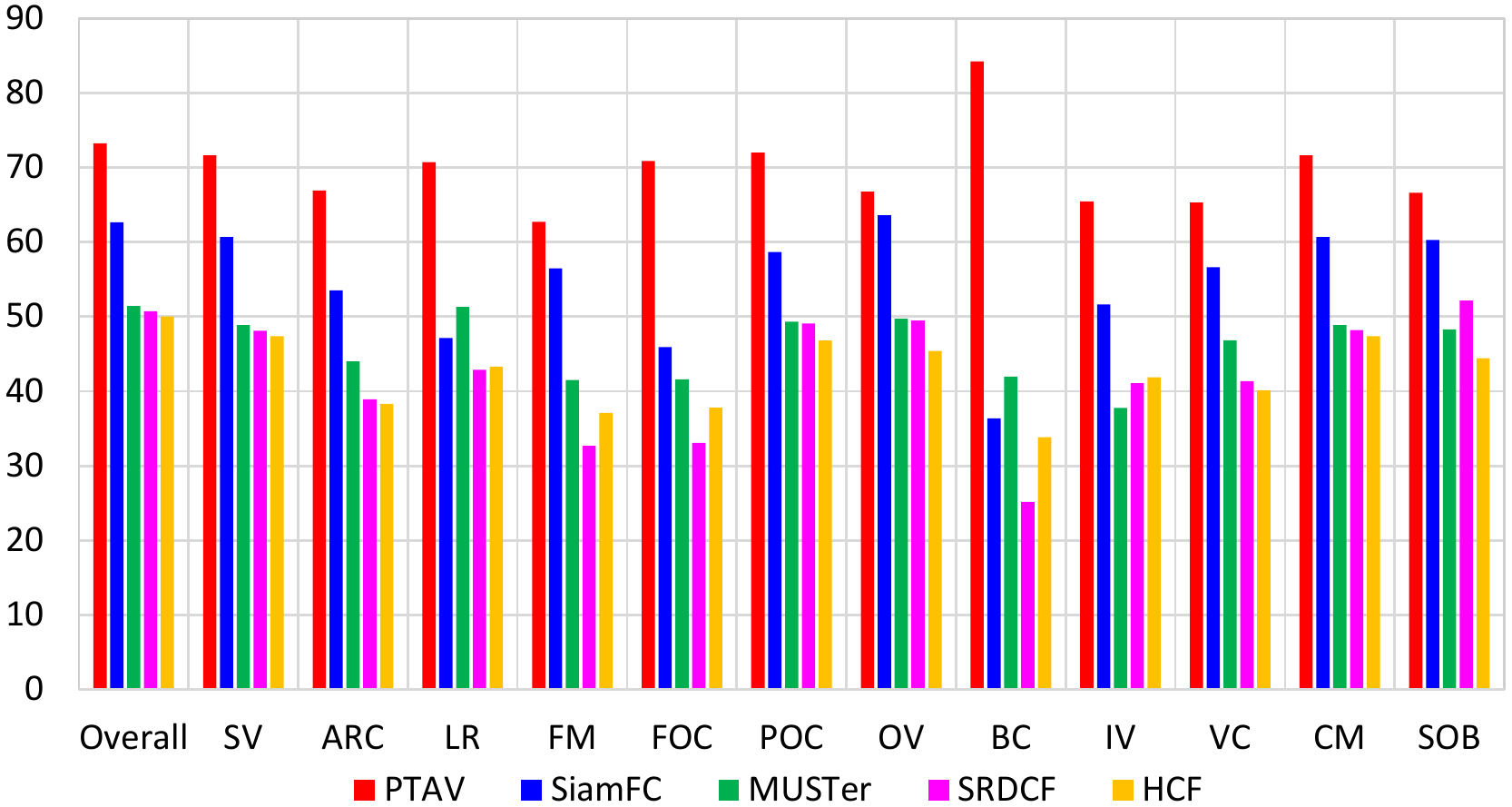}
	\caption{Average DPR (\%) in term of attributes on UAV20L~\cite{mueller2016benchmark}.}
	\label{UAL20L_att}
\end{figure}

Following~\cite{mueller2016benchmark}, we report evaluation results in Figure \ref{comparison_UAL}. PTAV achieves the best performance in both DPR (73.2\%) and OSR (50.4\%), outperforming other approaces by large margins (6.2\% and 10.1\% compared with the second best in DPR and OSR, respectively). Furthermore, we analyze the performance of PTAV on twelve individual attributes provided with UAV20L~\cite{mueller2016benchmark}, icnluding scale variation (SV), aspect ratio change (ARC), low resolution (LR), fast motion (FM), full occlusion (FOC), partial occlusion (POC), out-of-view (OV), background clutter (BC), illumination variation (IV), viewpoint change (VC), camera motion (CM) and similar object (SOB). Figure \ref{UAL20L_att} displays the average DPR for five trackers on different attributes, and PTAV achieves the best results on each attribute.

\subsection{Experiment on VOT2016}

Finally, we test PTAV on the VOT2016 challenge~\cite{kristan2016visual}, which contains 60 challenging sequences. VOT2016 aims at evaluating short-term tracking performance and thus a tracker is re-initialized whenever failure happens. In other words, a tracker is reset if its tracking results are found unreliable. Nevertheless, this protocol is not directly applicable to our tracker since PTAV automatically detects failures by itself and rolls back to resume tracking.

To follow the above evaluation protocol, we modify PTAV by running it multiple rounds with different starting frames. In particular, in each round, we run PTAV at current starting frame without resetting. For the first round, the first frame in the input sequence is used as the start frame. Afterward, we compare the tracking results with groundtruth to find the first failure using the VOT2016 protocol, and then we re-initialize PTAV from the failure frame for the next round. We repeat this process until no failure is detected. On VOT2016, PTAV runs at 25 \emph{fps}.

\begin{table}[!t]
	\centering
	\caption{Comparisons with state-of-the-art tracking methods on VOT2016~\cite{kristan2016visual} in terms of expected average overlap (EAO\%), accuracy (\%), robustness (\%) and no-reset average overlap (AO\%). The best two results are highlighted in \textcolor{red}{red} and \textcolor{blue}{blue} fonts, respectively.}
	\begin{tabular}{rcccc}
		\hline
		Algorithms & EAO   & Accuracy  & Robustness  & AO \\
		\hline\hline
		PTAV (Ours)  & 31.2 & \textcolor{blue}{56.1} & 27.9 & 43.2 \\
		\hline
		C-COT~\cite{danelljan2016beyond} & \textcolor{red}{33.1} & 53.9 & \textcolor{blue}{23.8} & 46.9 \\
		TCNN~\cite{nam2016modeling}  & \textcolor{blue}{32.5} & 55.4 & 26.8 & \textcolor{blue}{48.5} \\
		SSAT~\cite{kristan2016visual}  & 32.1 & \textcolor{red}{57.7} & 29.1 & \textcolor{red}{51.5} \\
		MLDF~\cite{kristan2016visual}  & 31.1 & 49.0 & \textcolor{red}{23.3} & 42.8 \\
		Staple~\cite{bertinetto2016staple} & 29.5 & 54.4 & 37.8 & 38.8 \\
		DDC~\cite{kristan2016visual}   & 29.3 & 54.1 & 34.5 & 39.1 \\
		EBT~\cite{zhu2016beyond}   & 29.1 & 46.5 & 25.2 & 37.0 \\
		SRBT~\cite{kristan2016visual}  & 29.0 & 49.6 & 35.0 & 33.3 \\
		Staple+~\cite{kristan2016visual} & 28.6 & 55.7 & 36.8 & 39.2 \\
		DNT~\cite{chi2017dual}   & 27.8 & 51.4 & 32.9 & 42.7 \\
		\hline
	\end{tabular}%
	\label{vot16}%
\end{table}%

PTAV is compared with top ten trackers in the VOT2016 challenge, including C-COT~\cite{danelljan2016beyond}, TCNN~\cite{nam2016modeling}, SSAT~\cite{kristan2016visual}, MLDF~\cite{kristan2016visual}, Staple~\cite{bertinetto2016staple}, DDC~\cite{kristan2016visual}, EBT~\cite{zhu2016beyond}, SRBT~\cite{kristan2016visual}, Staple+~\cite{kristan2016visual} and DNT~\cite{chi2017dual}. Table \ref{vot16} demonstrates comparison results in VOT2016. It shows that C-COT~\cite{danelljan2016beyond} and TCNN~\cite{nam2016modeling} achieve the best results with EAOs of 33.1\% and 32.5\%, respectively. C-COT~\cite{danelljan2016beyond} utilizes deep features to model object appearance and TCNN~\cite{nam2016modeling} proposes tree-structured CNNs for tracking with online update. Despite obtaining superior performances, their speeds are around 1 and 2 \emph{fps}. By contrast, PTAV achieves competitive result (EAO of 31.2\%), while running in real-time.

\subsection{Ablation Study}

\renewcommand\arraystretch{1}
\begin{table*}[!t]
  \centering
  \caption{Comparisons of DPR (\%), OSR (\%), CLE in pixels and speed (fps) among different $\TK$ with VGGNet~\cite{simonyan2014very} based $\VRF$ on three benchmarks.}
    \begin{tabular}{@{}R{1.8cm}|@{}C{1.1cm}@{}C{1.1cm}@{}C{1.1cm}@{}C{1.1cm}|@{}C{1.1cm}@{}C{1.1cm}@{}C{1.1cm}@{}C{1.1cm}|@{}C{1.1cm}@{}C{1.1cm}@{}C{1.1cm}@{}C{1.1cm}}
    \hline
          & \multicolumn{4}{c|}{OTB2015~\cite{wu2015object}}    & \multicolumn{4}{c|}{TC128~\cite{liang2015encoding}}     & \multicolumn{4}{c}{UAV20L~\cite{mueller2016benchmark}} \\
    \cline{2-13}          & DPR   & OSR   & CLE   & Speed & DPR   & OSR   & CLE   & Speed & DPR  & OSR   & CLE   & Speed \\
    \hline\hline
    PTAV$_{\mathrm{Staple}}$  & 86.2  & 77.9  & 18.9     & 27    & 77.2  & 70.0  & 32.1     & 23    & 73.2  & 62.4  & 56.1     & 30 \\
    PTAV$_{\mathrm{fDSST}}$  & 85.2  & 77.9  & 19.4     & 27    & 75.2  & 66.1  & 32.7     & 24    & 63.1  & 47.8  & 102.7     & 26 \\
    PTAV$_{\mathrm{KCF}}$  & 74.2  & 58.6  & 34.5     & 24    & 63.9  & 54.6  & 53.6     & 19    & 41.6  & 32.1  & 195.3     & 20 \\
    \hline
    Staple~\cite{bertinetto2016staple} & 78.4  & 70.9  & 31.9     & 43    & 66.7  & 62    & 57.5     & 43    & 48.5  & 44.3  & 223.1     & 49 \\
    fDSST~\cite{danelljan2016discriminative} & 72.0  & 67.6  & 51.1     & 51    & 57.5  & 52.4  & 82.1     & 51    & 42.2  & 34.4  & 256.8     & 52 \\
    KCF~\cite{henriques2015high}   & 69.2  & 54.8  & 45.0     & 243   & 55.1  & 46.1  & 77.4     & 242   & 31.1  & 20.0  & 282.4     & 245 \\
    \hline
    \end{tabular}%
  \label{tab:compare_t}%
\end{table*}%

\renewcommand\arraystretch{1}
\begin{table*}[!t]
  \centering
  \caption{Comparisons of DPR (\%), OSR (\%), CLE in pixels and speed (fps) between different $\VRF$ with same tracker on three benchmarks.}
    \begin{tabular}{c|c|@{}C{1.1cm}@{}C{1.1cm}@{}C{1.1cm}@{}C{1.1cm}|@{}C{1.1cm}@{}C{1.1cm}@{}C{1.1cm}@{}C{1.1cm}|@{}C{1.1cm}@{}C{1.1cm}@{}C{1.1cm}@{}C{1.1cm}}
    \hline
    \multicolumn{1}{c}{} &       & \multicolumn{4}{c|}{OTB2015~\cite{wu2015object}}    & \multicolumn{4}{c|}{TC128~\cite{liang2015encoding}}     & \multicolumn{4}{c}{UAV20L~\cite{mueller2016benchmark}} \\
    \cline{3-14}    \multicolumn{1}{c}{} &       & DPR   & OSR   & CLE   & Speed & DPR   & OSR   & CLE   & Speed & DPR   & OSR   & CLE   & Speed \\
    \hline\hline
    \multirow{2}[0]{*}{Staple} & PTAV$_\mathrm{VGGNet}$  & 86.2  & 77.9  &   18.9    & 27    & 77.2  & 70    &    32.1   & 23    & 73.2  & 62.4  & 56.1      & 30 \\
          & PTAV$_\mathrm{AlexNet}$  & 84.0    & 75.5  &   20.7    & 34    & 75.0    & 68.9  &   40.3    & 31    & 65.9  & 58.6  &   70.7    & 33 \\
    \hline
    \multirow{2}[0]{*}{fDSST} & PTAV$_\mathrm{VGGNet}$  & 85.2  & 77.9  &   19.4    & 27    & 75.2  & 66.1  &    32.7   & 24    & 63.1  & 47.8  & 102.7      & 26 \\
          & PTAV$_\mathrm{AlexNet}$  & 79.4  & 74.3  &    31.2   & 29    & 67.5  & 61.1  &   43.8    & 26    & 54.7  & 43.2  &   121.0    & 31 \\
    \hline
    \multirow{2}[0]{*}{KCF} & PTAV$_\mathrm{VGGNet}$  & 74.2  & 58.6  &     34.5  & 24    & 63.9  & 54.6  &    53.6   & 19    & 41.6  & 32.1  &  195.3     & 20 \\
          & PTAV$_\mathrm{AlexNet}$  & 72.3  & 58.0    &    37.1   & 31    & 62.4  & 52.6  &   54.9    & 22    & 39.6  & 27.6  &   216.1    & 27 \\
    \hline
    \end{tabular}%
  \label{tab:compare_v}%
\end{table*}%

\renewcommand\arraystretch{1}
\begin{table*}[!t]
  \centering
  \caption{Comparisons of DPR (\%), OSR (\%), CLE in pixels and speed (fps) between fixed template and dynamic template set using $\VRF$ based on Staple~\cite{bertinetto2016staple} and $\TK$ based on VGGNet~\cite{simonyan2014very} on three benchmarks.}
  \begin{tabular}{C{2.5cm}|@{}C{1.1cm}@{}C{1.1cm}@{}C{1.1cm}@{}C{1.1cm}|@{}C{1.1cm}@{}C{1.1cm}@{}C{1.1cm}@{}C{1.1cm}|@{}C{1.1cm}@{}C{1.1cm}@{}C{1.1cm}@{}C{1.1cm}}
  \hline
  & \multicolumn{4}{c|}{OTB2015~\cite{wu2015object}}   & \multicolumn{4}{c|}{TC128~\cite{liang2015encoding}}    & \multicolumn{4}{c}{UAV20L~\cite{mueller2016benchmark}} \\
  \cline{2-13} & DPR & OSR & CLE & Speed & DPR & OSR & CLE & Speed & DPR & OSR & CLE & Speed \\
  \hline\hline
  dynamic templates & 86.2  & 77.9  & 18.9  & 27.0  & 77.2  & 70.0  & 32.1  & 23.0  & 73.2  & 62.4  & 56.1  & 30.0 \\
  fixed template & 85.6  & 77.1  & 20.4  & 25.0  & 76.3  & 68.9  & 32.3  & 22.0  & 72.6  & 61.6  & 62.8  & 28.0 \\
  \hline
  \end{tabular}%
  \label{tab:comparison_template}%
\end{table*}%

\subsubsection{Different trackers for $\TK$}

In PTAV, $\TK$ is required to be efficient and accurate most of the time. To demonstrate the effects of different $\TK$, we compare three different base tracking algorithms including Staple~\cite{bertinetto2016staple} (the choice in this paper), fDSST~\cite{danelljan2016discriminative} and KCF~\cite{henriques2015high}. Among these trackers, KCF~\cite{henriques2015high} runs the most efficiently while least accurately in short time. Compared with KCF~\cite{henriques2015high} and fDSST~\cite{danelljan2016discriminative}, Staple~\cite{bertinetto2016staple} performs more robustly since it utilizes color information for tracking, which results in its relative inefficiency. The comparison results on OTB2015~\cite{wu2015object}, TC128~\cite{liang2015encoding} and UAV20L~\cite{mueller2016benchmark} are shown in Table \ref{tab:compare_t}.

From Table \ref{tab:compare_t}, we can see that PTAV with the Staple base tracker (PTAV$_{\mathrm{Staple}}$) performs better than those with fDSST (PTAV$_{\mathrm{fDSST}}$) and KCF (PTAV$_{\mathrm{KCF}}$). Though KCF runs the fastest among these trackers, it performs least accurately in short time, resulting in more requests for verifications and detections, and significantly increased computations. As shown in Table \ref{tab:compare_t}, the speeds of PTAV$_{\mathrm{KCF}}$ on OTB2015~\cite{wu2015object}, TC128~\cite{liang2015encoding} and UAV20L~\cite{mueller2016benchmark} are respectively 24, 19 and 20 \emph{fps}, which are much slower than PTAV$_{\mathrm{Staple}}$ (27, 23 and 30 \emph{fps}, respectively) and PTAV$_{\mathrm{fDSST}}$ (27, 24 and 26 \emph{fps}, respectively). 

In terms of tracking accuracy, on OTB2015~\cite{wu2015object}, PTAV$_{\mathrm{fDSST}}$ achieves competitive performance (85.2\% of DPR and 77.9 \% of OSR) compared to PTAV$_{\mathrm{Staple}}$ (86.2\% of DPR and 77.9\% of OSR). However, on the more challenging UAV20L~\cite{mueller2016benchmark}, PTAV$_{\mathrm{Staple}}$ significantly outperforms PTAV$_{\mathrm{fDSST}}$ in both accuracy and efficiency. Specifically, PTAV$_{\mathrm{Staple}}$ obtains a DPR of 73.2\%, an OSR of 62.4\% and speed of 30 \emph{fps} while PTAV$_{\mathrm{fDSST}}$ with DPR of 63.1\%, OSR of 47.8\% and speed of 26 fps. The main reason accounting for this is that the baseline Staple leverages color cues for tracking. In UAV20L~\cite{mueller2016benchmark}, the tracking target frequently suffers from severe view changes, which are fatal to HoG features. Nevertheless, Staple is able to deal with view changes using color statistics, and thus performs better than fDSST in short periods. As a consequence, PTAV$_{\mathrm{Staple}}$ performs more favorably than PTAV$_{\mathrm{fDSST}}$ and requires less verifications and detections, further improving efficiency. Besides, we observe from Table \ref{tab:compare_t} that all the three PTAV versions improve their baseline trackers by large margins.

\subsubsection{Different verifiers for $\VRF$}
The verifier $\VRF$ plays a crucial role in PTAV by validating tracking results and correcting $\TK$ if needed. To guarantee the quality of verification, $\VRF$ is required to be as accurate as possible. To such purpose, we adopt the Siamese networks~\cite{chopra2005learning} for $\VRF$. To study the effects of different $\VRF$, we compare two different alternatives: one is based on VGGNet~\cite{simonyan2014very} in previous experiments, and the other utilizes the much lighter AlexNet~\cite{krizhevsky2012imagenet}\footnote{For the verifier based on AlexNet~\cite{krizhevsky2012imagenet}, one just needs to replace and initialize the five convolutional blocks in Figure \ref{siamese_net} with AlexNet.}. Compared to the VGGNet-based $\VRF$, the AlexNet-based $\VRF$ runs more efficiently but less accurately. Specifically, the speed of VGGNet based-$\VRF$ is 6 \emph{fps} while while its AlexNet counterpart runs at 17 fps. The comparison results of different verifiers with the same tracker on OTB2015, TC128 and UAV20L are reported in Table \ref{tab:compare_v}.

From Table \ref{tab:compare_v}, we observe that, when using the same base tracker, PTAV with VGGNet-based $\VRF$ (PTAV$_\mathrm{VGGNet}$) outperforms that with AlexNet-based $\VRF$ (PTAV$_\mathrm{AlexNet}$) on all three benchmarks. For example, by selecting Staple as the tracking part, PTAV$_\mathrm{VGGNet}$ achieves DPRs of 86.2\%, 77.2\% and 73.2\% on OTB2015, TC128 and UAV20L, respectively, obtaining improvements of respectively 2.2\%, 2.2\% and 7.3\% compared to PTAV$_\mathrm{AlexNet}$ with DPRs of 84.0\%, 75.0\% and 65.9\%, respectively. In PTAV, an accurate $\VRF$ is able to effectively reduce detections, which in turn decreases computation for verifications and leads to running time efficiency. Consequently, though AlexNet is computationally much more efficient than VGGNet, the speed of PTAV$_\mathrm{VGGNet}$ is competitive to that of PTAV$_\mathrm{AlexNet}$, and it runs in real-time (27, 23 and 30 \emph{fps} on OTB2015, TC128 and UAV20L, respectively), as shown in Table \ref{tab:compare_v}.

\renewcommand\arraystretch{1.1}
\begin{table}[!t]\footnotesize
	\centering
	\caption{Comparisons of different $N_{\mathrm{int}}$ in DPR and speed on OTB2015~\cite{wu2015object}.}
	\begin{tabular}{rccc}
		\hline
		&  $N_{\mathrm{int}}=5$ & $N_{\mathrm{int}}=10$ & $N_{\mathrm{int}}=15$ \\
		\hline\hline
		DPR (\%) & 86.3    & 86.2    & 84.6 \\
		\hline
		Speed (fps) & 25    & 27    & 30 \\
		\hline
	\end{tabular}%
	\label{differentV}%
\end{table}%

\renewcommand\arraystretch{1.1}
\begin{table}[!t]
	\centering
	\caption{Comparisons of tracking speed (fps) on three benchmarks.}
	\begin{tabular}{@{}R{1.65cm}@{}C{1.85cm}@{}C{1.55cm}@{}@{}C{1.65cm}@{}C{1.9cm}@{}}
		\hline
		&  OTB2015~\cite{wu2015object} & TC128~\cite{liang2015encoding} &UAV20L~\cite{mueller2016benchmark} & VOT2016~\cite{kristan2016visual}\\
		\hline\hline
		Single thread &  15    & 14 & 17 & 15\\
		\hline
		Two threads &  27    & 23 & 30 & 25 \\
		\hline
	\end{tabular}%
	\label{speed}%
\end{table}%

\subsubsection{Fixed template v.s. dynamic template set}

To adapt $\VRF$ to target appearance variation, we propose the dynamic template set for adaptive verification, which can take advantages of confident tracking results to improve validation quality, leading to reduction of verifications and detections for efficiency. Table \ref{tab:comparison_template} shows the comparison results of PTAV using dynamic template set versus using a fixed template. From Table \ref{tab:comparison_template}, we observe that using dynamic template set for verification improves PTAV in both accuracy and efficiency. In specific, the DPRs on three benchmarks are improved from 85.6\%, 76.3\% and 72.6\% to 86.2\%, 77.2\% and 73.2\%, respectively. The speeds are boosted from 25, 22 and 28 \emph{fps} to 27, 23 and 30 \emph{fps}, respectively.

\subsubsection{Different verification interval $N_{\mathrm{int}}$}

In PTAV, different verification interval $N_{\mathrm{int}}$ may affect both the accuracy and efficiency. A smaller $N_{\mathrm{int}}$ implies more frequent verification, which requires more computation and thus degrades the efficiency. A larger $N_{\mathrm{int}}$, on the contrary, may cost less computation but may put PTAV at the risk when the target appearance changes quickly. If the tracker loses the tracking target, it may update vast backgrounds in its appearance model until next verification. Even if the verifier re-locates the target and offers a correct detection result, the tracker may still lose it due to heavy changes in the target appearance model. Table \ref{differentV} demonstrates sampled results with three different $N_{\mathrm{int}}$ on OTB2015~\cite{wu2015object}. Taking into account both accuracy and speed, we set $N_{\mathrm{int}}$ to 10 in our experiments.

\subsubsection{Two threads v.s. one}

In PTAV, the tracker $\TK$ does not rely on the verifier $\VRF$  most of the time, and two separate threads process tracking and verifying in parallel for efficiency. Consequently, $\TK$ does not have to wait for the feedback from $\VRF$ to process next frame, and it traces back and resumes tracking only when receiving a correction feedback from $\VRF$. Owing to storing intermediate results, $\TK$ is able to quickly trace back without little extra computation. Table \ref{speed} shows the comparison of speed between using two threads and using only a single thread. It shows that using two threads in parallel clearly improves the efficiency of the system.

\subsection{Failure Cases}

\begin{figure}[!t]
\centering
\includegraphics[width=.24\linewidth,height=.14\linewidth]{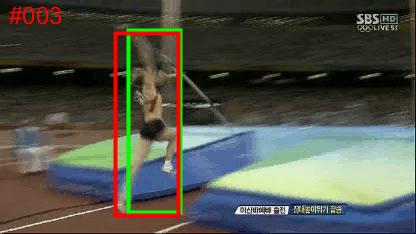} \includegraphics[width=.24\linewidth,height=.14\linewidth]{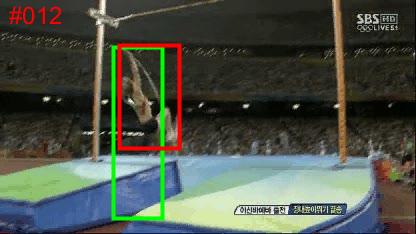} \includegraphics[width=.24\linewidth,height=.14\linewidth]{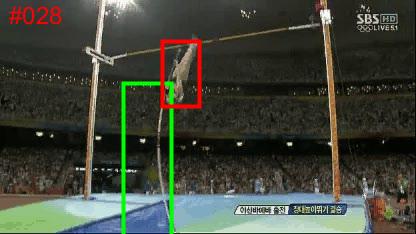} \includegraphics[width=.24\linewidth,height=.14\linewidth]{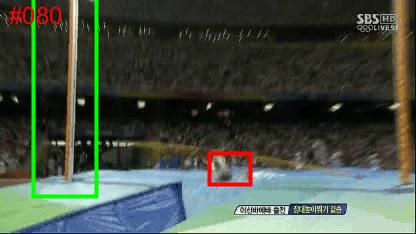}\\
\vspace{1mm}\includegraphics[width=.24\linewidth,height=.14\linewidth]{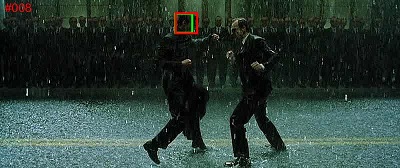} \includegraphics[width=.24\linewidth,height=.14\linewidth]{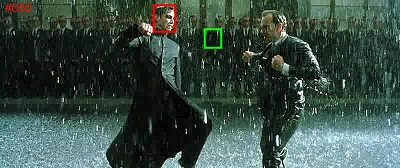} \includegraphics[width=.24\linewidth,height=.14\linewidth]{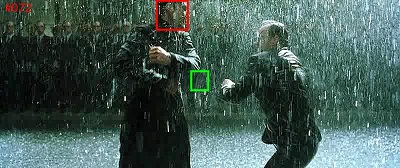} \includegraphics[width=.24\linewidth,height=.14\linewidth]{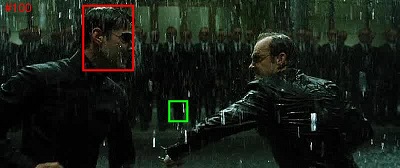} \\
\includegraphics[width=3.5cm]{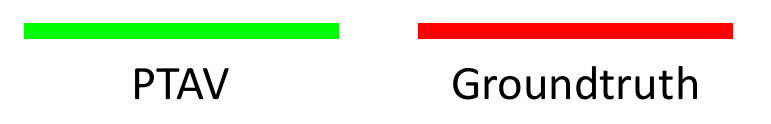}
\caption{Failure cases of \emph{Jump} and \emph{Matrix} for PTAV on OTB2015~\cite{wu2015object}.}
\label{failure}
\end{figure}

With the collaboration between $\TK$ and $\VRF$, PTAV usually performs well in dealing with various challenging situations; however, there exist scenarios in which PTAV may fail. Shown in Figure \ref{failure}(a) on \emph{Jump}, though $\VRF$ can accurately detect unreliable tracking results of $\TK$, it does not provide correct feedbacks for subsequent tracking due to heavy deformation. On \emph{Matrix} shown in Figure \ref{failure}(b), the target undergoes severe illumination variation, occlusion, rotation and background cluttering, causing difficulties for $\TK$ to localize the target. Even though $\VRF$ corrects $\TK$ when validating unreliable tracking results, $\TK$ still drifts to background quickly. In certain cases where the target suffers form heavy appearance changes, $\VRF$ cannot provide effective feedbacks to $\TK$, resulting in irrecoverable drift and failures.

\section{Conclusion}
\label{con}
In this paper, we propose a new visual tracking framework, \emph{parallel tracking and verifying} (PTAV), which decomposes object tracking into two sub-tasks, fast tracking and reliable verifying. We show that, by carefully distributing the two tasks into two parallel threads and allowing them to work together, PTAV can achieve the best known tracking accuracy among all real-time tracking algorithms. Furthermore, to adapt the verifier to object appearance variations, we propose using dynamic target templates for adaptive verification, resulting in further improvements in both accuracy and efficiency. The encouraging results are demonstrated in extensive experiments on four popular benchmarks. Moreover, PTAV is a flexible framework with great rooms for improvement and generalization, and thus is expected to inspire the designing of more efficient tracking algorithms in the future.


\ifCLASSOPTIONcaptionsoff
  \newpage
\fi

\bibliographystyle{IEEEtran}
\bibliography{mybibfile}

\begin{IEEEbiography}[{\includegraphics[width=25.4 mm,height=31.75 mm]{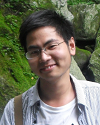}}]{Heng Fan}
received his B.E. degree in College of Science, Huazhong Agricultural University (HZAU), Wuhan, China, in 2013. He is currently a Ph.D. student in the Department of Computer and Information Science, Temple University, Philadelphia, USA. His research interests include computer vision, pattern recognition and machine learning.
\end{IEEEbiography}

\begin{IEEEbiography}[{\includegraphics[width=25.4 mm,height=30.75 mm]{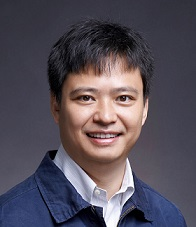}}]{Haibin Ling}
received the BS and MS degrees from Peking University, China, in 1997 and 2000,
respectively, and the PhD degree from the University of Maryland College Park in 2006. From 2000 to 2001, he was an assistant researcher at Microsoft Research Asia. From 2006 to 2007, he worked as a postdoctoral scientist at the University of California Los Angeles. After that, he joined Siemens Corporate Research as a research scientist. Since fall 2008, he has been with Temple University where he is now an Associate Professor.
Ling's research interests include computer vision, augmented reality, medical image analysis, and human computer interaction. He received the Best Student Paper Award at the ACM Symposium on User Interface Software and Technology (UIST) in 2003, and the NSF CAREER Award in 2014. He serves as associate editors for IEEE Trans. on Pattern Analysis and Machine Intelligence, Pattern Recognition, and Computer Vision and Image Understanding, and has served as area chairs for CVPR 2014 and CVPR 2016.
\end{IEEEbiography}


\end{document}